\documentclass[review,3p]{elsarticle}

\usepackage{subfigure}
\usepackage{amssymb}
\usepackage{amsmath}
\usepackage{multirow, booktabs}

\usepackage{lineno}
\usepackage{svg}
\usepackage{ulem}
\usepackage{adjustbox}

\usepackage{algorithm}
\usepackage{algorithmic}

\usepackage{color, soul}
\usepackage{hyperref}
\hypersetup{colorlinks=true,
            linkcolor=blue,
            anchorcolor=blue,
            citecolor=blue}

\journal{Pattern Recognition}

\begin{document}

\begin{frontmatter}



\title{Detect-Order-Construct: A Tree Construction based Approach for Hierarchical Document Structure Analysis}

\author[ustc,msra]{Jiawei Wang\corref{ca}\fnref{myfootnote}}
\ead{wangjiawei@mail.ustc.edu.cn}

\author[ustc,msra]{Kai Hu\corref{ca}\fnref{myfootnote}}
\ead{hk970213@mail.ustc.edu.cn}

\author[msra]{Zhuoyao Zhong\corref{ca}\fnref{myfootnote}}
\ead{zhuoyao.zhong@gmail.com}

\author[msra]{Lei Sun\fnref{myfootnote}}
\ead{kuangtongustc@gmail.com}

\author[msra]{Qiang Huo}
\ead{qianghuo@microsoft.com}

\affiliation[ustc]{organization={Department of EEIS, University of Science and Technology of China},
            city={Hefei},
            postcode={230026}, 
            country={China}}

\affiliation[msra]{organization={Microsoft Research Asia},
            city={Beijing},
            postcode={100080}, 
            country={China}}

\fntext[myfootnote]{Work done when Jiawei Wang and Kai Hu were interns, Zhuoyao Zhong and Lei Sun were FTEs at Multi-Modal Interaction Group, Microsoft Research Asia, Beijing, China.}
\cortext[ca]{Corresponding author.}

\begin{abstract}
Document structure analysis (aka document layout analysis) is crucial for understanding the physical layout and logical structure of documents, with applications in information retrieval, document summarization, knowledge extraction, etc. In this paper, we concentrate on Hierarchical Document Structure Analysis (HDSA) to explore hierarchical relationships within structured documents created using authoring software employing hierarchical schemas, such as LaTeX, Microsoft Word, and HTML. To comprehensively analyze hierarchical document structures, we propose a tree construction based approach that addresses multiple subtasks concurrently, including page object detection (Detect), reading order prediction of identified objects (Order), and the construction of intended hierarchical structure (Construct). We present an effective end-to-end solution based on this framework to demonstrate its performance. To assess our approach, we develop a comprehensive benchmark called Comp-HRDoc, which evaluates the above subtasks simultaneously. Our end-to-end system achieves state-of-the-art performance on two large-scale document layout analysis datasets (PubLayNet and DocLayNet), a high-quality hierarchical document structure reconstruction dataset (HRDoc), and our Comp-HRDoc benchmark. The Comp-HRDoc benchmark is publicly available at \url{https://github.com/microsoft/CompHRDoc}.

\end{abstract}



\begin{keyword}
Document Layout Analysis \sep Table of Contents \sep Reading Order Prediction \sep Page Object Detection


\end{keyword}

\end{frontmatter}


\section{Introduction}
Document Structure Analysis (DSA) is a comprehensive process that identifies the fundamental components within a document, encompassing headings, paragraphs, lists, tables, and figures, and subsequently establishes the logical relationships and structures of these components. This process results in a structured representation of the document's physical layout that accurately mirrors its logical structure, thereby enhancing the effectiveness and accessibility of information retrieval and processing. In a contemporary digital landscape, the majority of mainstream documents are structured creations, crafted using hierarchical-schema authoring software such as LaTeX, Microsoft Word, and HTML. Consequently, Hierarchical Document Structure Analysis (HDSA), which focuses on extracting and reconstructing the inherent hierarchical structures within these document layouts, has gained significant attention. However, despite its burgeoning popularity, HDSA poses a substantial challenge due to the diversity of document content and the intricate complexity of their layouts.

Over the past three decades, document structure analysis has garnered significant interest in the research community. Early research efforts primarily focused on physical layout analysis and logical structure analysis, employing various approaches such as knowledge-based \cite{kreich1991experimental}, rule-based \cite{tsujimoto1990understanding}, model-based \cite{yamashita1991model}, and grammar-based \cite{krishnamoorthy1993syntactic} methods. However, these traditional methods face limitations in terms of effectiveness and scalability due to their susceptibility to noise, ambiguity, and difficulties in handling complex document collections. Furthermore, the absence of quantitative performance evaluations hinders the proper evaluation of these techniques.
In the era of deep learning, a growing number of deep learning based approaches have been applied to the field of document structure analysis, leading to notable improvements in performance and robustness. However, these methods primarily focus on specific sub-tasks of DSA, such as Page Object Detection, Reading Order Prediction, and Table of Contents (TOC) Extraction, among others. Despite the substantial progress achieved in these individual sub-tasks, there remains a gap in the research community for a comprehensive end-to-end system or benchmark that addresses all aspects of document structure analysis concurrently. Filling this gap would significantly advance the field and encourage further research in this area. 

Recently, hierarchical document structure analysis has gained traction with representative explorations like DocParser and HRDoc. DocParser \cite{rausch2021docparser} is the an end-to-end system for parsing document renderings into hierarchical document structures, encompassing all text elements, nested figures, tables, and table cell structures. Initially, the system employs Mask R-CNN \cite{he2017mask} to detect all document entities within a document image. Subsequently, it devises a set of rules to predict two predefined relationships (i.e., ``parent\_of" and ``followed\_by") between document entities to parse the complete physical structure of the document. However, the system does not take into account the logical structure of documents, such as the table of contents, and its reliance on a rule-based approach considerably limits its overall effectiveness and adaptability. On the other hand, HRDoc \cite{Ma2023HRDoc} proposed an encoder-decoder based hierarchical document structure parsing system (DSPS) to reconstruct the hierarchical structure of documents. This system employs a multi-modal bidirectional encoder and a structure-aware GRU decoder to predict the logical roles of the text-lines and the relationships between them. Although DSPS achieves significant performance improvements over DocParser and considers the logical structure of documents, it presumes that the reading order of the document is provided, which is an essential aspect of document structure analysis that should not be overlooked. Furthermore, with the increase in text-lines within documents, the computational complexity of DSPS grows quadratically, presenting significant challenges when processing longer documents. Additionally, predicting relationships between line-level semantic units may result in the loss of broader contextual information, which is crucial for a comprehensive understanding of the document's structure.

\begin{figure}[t]
    \centering
    \includegraphics[width=0.9\linewidth]{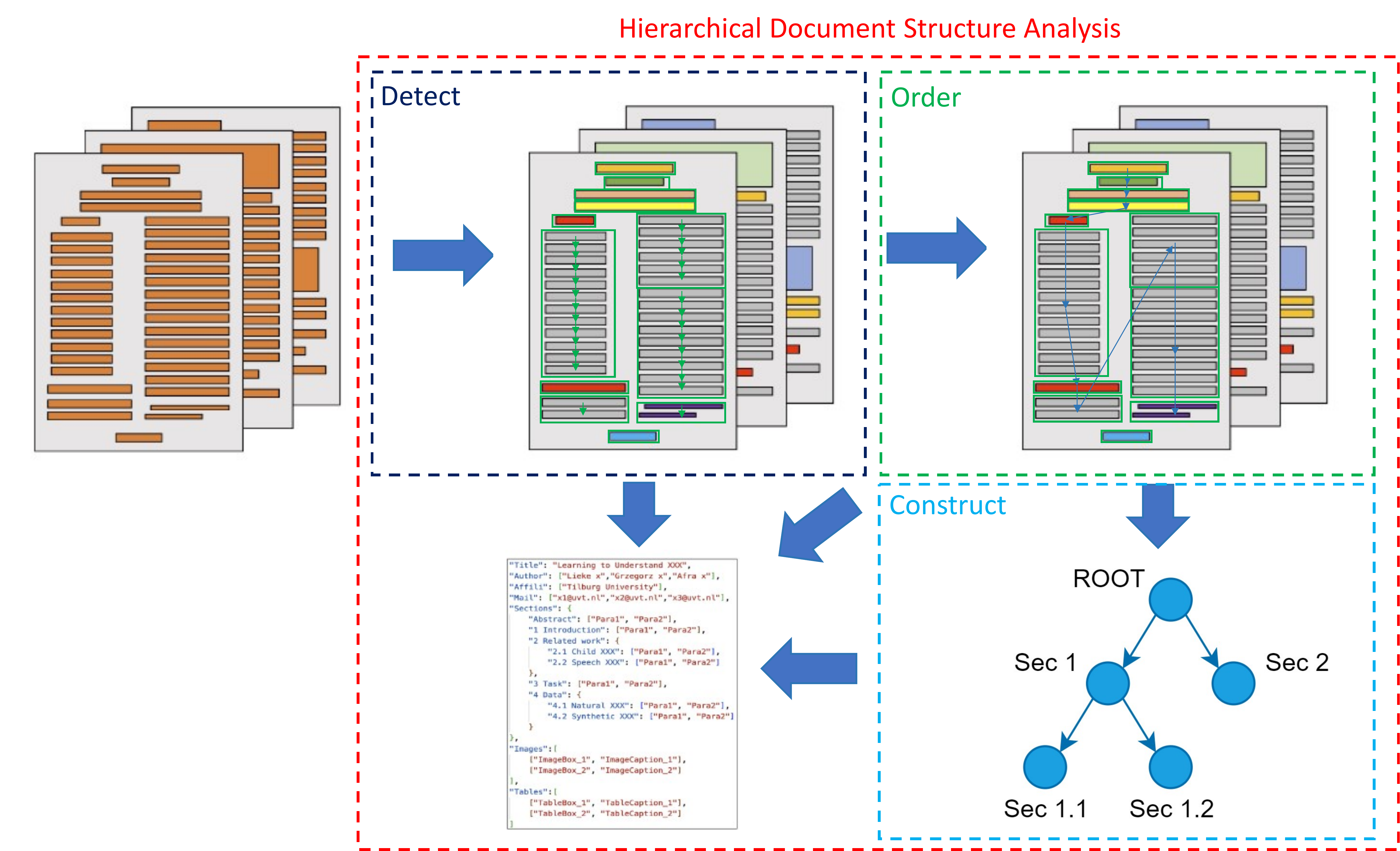}
    \caption{Overview of our tree construction based approach, named Detect-Order-Construct, for hierarchical document structure analysis.}
    \label{fig:pipeline}
\end{figure}

In this study, we propose a comprehensive approach to thoroughly analyzing hierarchical document structures using a tree construction based method. This method decomposes tree construction into three distinct stages, namely Detect, Order, and Construct, as illustrated in Fig.~\ref{fig:pipeline}. Initially, given a set of document images, the Detect stage is dedicated to identifying all page objects and assigning a logical role to each object, thereby forming the nodes of the hierarchical document structure tree. Following this, the Order stage establishes the reading order relationships among these nodes, which corresponds to a pre-order traversal of the hierarchical document structure tree. Finally, the Construct stage identifies hierarchical relationships (e.g., Table of Contents) between semantic units to construct an abstract hierarchical document structure tree. By integrating the results of all three stages, we can effectively construct a complete hierarchical document structure tree, facilitating a more comprehensive understanding of complex documents. 

To demonstrate its performance, we present an effective end-to-end solution based on this framework. For the Detect stage, we consider OCR'd text-lines as the basic semantic units and introduce a novel hybrid method, which combines a top-down model with a relation prediction model to simultaneously detect graphical page objects (e.g., tables, figures, etc.), group text-lines into text regions according to the intra-region reading order, and recognize the logical roles of text regions. Any top-down object detection or instance segmentation models can be directly applied to detecting graphical page objects, sharing a visual backbone network with the relation prediction model. Subsequently, we can cohesively formalize these three stages as relation prediction tasks by defining distinct types of relationships. They include the intra-region reading order relationships between text-lines to group text-lines into text regions, inter-region reading order relationships between text regions to generate the reading sequence of text regions, and TOC relationships between section headings to summarize the overall hierarchical document structure. To address these stages in a unified manner, we introduce a type of multi-modal transformer-based relation prediction models, which are designed to tackle all three stages. This novel relation prediction model approaches the relation prediction as a dependency parsing task, employing a multi-modal transformer encoder to model the interactions between input pairs via a global self-attention mechanism. Moreover, in response to the chain structure of reading order and the tree structure of table of contents, we design two structure-aware relation prediction models specifically tailored for these two structures, ensuring a more accurate and efficient analysis of these hierarchical relationships. 

Throughout these three stages, several sub-tasks play an integral role in hierarchical document structure analysis. Consequently, during the performance evaluation phase, it is not sufficient to merely assess the overall accuracy of hierarchical document structure reconstruction, as done in HRDoc \cite{Ma2023HRDoc}. An exhaustive and thorough evaluation of each sub-task involved is equally important. Leveraging the HRDoc dataset, we establish a comprehensive benchmark, Comp-HRDoc, aimed at evaluating page object detection, reading order prediction, table of contents extraction, and hierarchical structure reconstruction concurrently. Extensive experimental results demonstrate that our proposed end-to-end system achieves state-of-the-art performance on two large-scale document layout analysis datasets (i.e., PubLayNet \cite{zhong2019publaynet} and DocLayNet \cite{pfitzmann2022doclaynet}), and a hierarchical document structure reconstruction dataset (i.e., HRDoc). Moreover, our proposed comprehensive benchmark, Comp-HRDoc, effectively illustrate the effectiveness and superiority of our approach across all sub-tasks.

The main contributions of this paper are as follows:
\begin{itemize}

\item Proposed a tree construction based approach, namely Detect-Order-Construct, for hierarchical document structure analysis. To exemplify the effectiveness of this framework, we devise an effective end-to-end solution by casting uniformly the three-stage tasks as relation prediction problems. Furthermore, we design multi-modal transformer-based relation prediction models with two structure-aware improvements for chain structures and tree structures respectively to enhance the overall system performance.

\item Designed and established the first comprehensive benchmark, namely Comp-HRDoc, for the simultaneous evaluation of page object detection, reading order prediction, table of contents extraction, and hierarchical structure reconstruction.

\item Our proposed end-to-end system achieves state-of-the-art performance on two large-scale document layout analysis datasets (i.e., PubLayNet and DocLayNet), a hierarchical document structure reconstruction dataset (i.e., HRDoc) and our comprehensive benchmark Comp-HRDoc.
\end{itemize}

Although a preliminary study of the Detect stage in our end-to-end system has been presented in our conference paper \cite{zhong2023hybrid}, this paper significantly extends it in the following aspects: (1) A tree construction based approach, namely Detect-Order-Construct, is proposed for hierarchical document structure analysis; (2) A comprehensive benchmark is designed and established to simultaneously evaluate page object detection, reading order prediction, table of contents extraction, and hierarchical structure reconstruction; (3) Experimental results on a public benchmark dataset HRDoc \cite{Ma2023HRDoc} and our new benchmark Comp-HRDoc are presented to compare our approach with other works more comprehensively.
\section{Related Work}

Since the 1980s, numerous studies have been conducted on document structure analysis, which can be categorized into physical structure analysis (or physical layout analysis) and logical structure analysis \cite{mao2003survey}. Physical layout analysis focuses on identifying homogeneous regions of interest, also known as page objects, while logical structure analysis aims to assign logical roles to these regions and determine their relationships.
Early approaches to document structure analysis, mainly based on heuristic rules or grammar analysis, can be found in surveys \cite{tang1996automatic, mao2003survey}. In the past decade, a growing body of research \cite{zhong2019publaynet, yang2017learning, li2020docbank, pfitzmann2022doclaynet} has focused on document layout analysis, specifically physical layout analysis and logical role classification, which is also known as page object detection \cite{gao2017pod}. To maintain clarity, we will consistently use the term \textbf{``page object detection"} throughout this article to refer to the document layout analysis task that incorporates both physical layout analysis and logical role classification.
In addition to detecting page objects, numerous research studies have delved into the logical relationships between components within documents. These investigations have focused on aspects such as the reading order relationships and the organization of tables of contents. In this section, we primarily review and analyze recent developments in page object detection, reading order prediction, and hierarchical document structure reconstruction, providing an overview of the latest advancements and methodologies in these areas.

\subsection{Page Object Detection}
Page Object Detection, also known as POD \cite{gao2017pod}, is a task that involves locating logical objects (e.g., paragraphs, tables, mathematical equations, graphics, and figures) within document pages. Deep learning-based POD approaches can be broadly classified into three categories: object detection-based methods, semantic segmentation-based methods, and graph-based methods.

\textbf{Object detection-based methods.}
These methods leverage the latest top-down object detection or instance segmentation frameworks to address the page object detection problem. Pioneering efforts by Yi et al. \cite{yi2017cnn} and Oliveira et al. \cite{augusto2017fast} adapted R-CNN \cite{girshick2014rich} to identify and recognize page objects from document images. However, their performance was hindered by the limitations of traditional region proposal generation strategies. Subsequent research explored more sophisticated object detectors, such as Fast R-CNN \cite{girshick2015fast}, Faster R-CNN \cite{ren2015faster}, Mask R-CNN \cite{he2017mask}, Cascade R-CNN \cite{cai2019cascade}, SOLOv2 \cite{wang2020solov2}, CondInst \cite{tian2020conditional}, YOLOv5 \cite{yolov5}, and Deformable DETR \cite{deformdetr2021} as investigated by Vo et al. \cite{vo2018ensemble}, Zhong et al. \cite{zhong2019publaynet}, Saha et al. \cite{saha2019graphical}, Li et al. \cite{li2022dit}, Biswas et al. \cite{biswas2022docsegtr}, Hu et al. \cite{HU2024110212}, Pfitzmann et al. \cite{pfitzmann2022doclaynet}, and Yang et al. \cite{yang2022transformer}, respectively.
In addition, researchers have proposed effective techniques to enhance the performance of these detectors. For example, Zhang et al. \cite{zhang2021vsr} introduced a multi-modal Faster/Mask R-CNN model for page object detection that fused visual feature maps extracted by CNN with two 2D text embedding maps containing sentence and character embeddings. They also incorporated a graph neural network (GNN) based relation module to model the interactions between page object candidates. Shi et al. \cite{shi2022lateral} proposed a novel lateral feature enhancement backbone network, while Yang et al. \cite{yang2022transformer} employed Swin Transformer \cite{liu2021swin} as a more robust backbone network to boost the performance of Mask R-CNN and Deformable DETR for page object detection.
Recently, Gu et al. \cite{gu2022unified}, Li et al. \cite{li2022dit}, and Huang et al. \cite{huang2022layoutlmv3} further improved the performance of Faster R-CNN, Mask R-CNN, and Cascade R-CNN-based page object detectors by pre-training the vision backbone networks on large-scale document images using self-supervised learning algorithms. Despite achieving state-of-the-art results on several benchmark datasets, these methods continue to face challenges in detecting small-scale text regions.

\textbf{Semantic segmentation based methods.} 
These methods, such as those proposed by Yang et al. \cite{yang2017learning}, He et al. \cite{he2017multi}, Li et al. \cite{li2018deeplayout,li2019instance}, and Sang et al. \cite{sang2022exploiting}, typically employ existing semantic segmentation frameworks, such as FCN \cite{long2015fully}, to initially generate a pixel-level segmentation mask. Subsequently, the pixels are merged to form distinct types of page objects. Yang et al. \cite{yang2017learning} introduced a multi-modal FCN for page object segmentation, which combined visual feature maps and 2D text embedding maps with sentence embeddings to enhance pixel-wise classification accuracy. He et al. \cite{he2017multi} developed a multi-scale, multi-task FCN designed to concurrently predict a region segmentation mask and a contour segmentation mask. After refinement using a conditional random field (CRF) model, these two segmentation masks are processed by a post-processing module to obtain the final prediction results. Li et al. \cite{li2019instance} integrated label pyramids and deep watershed transformation into the vanilla FCN structure to prevent the merging of adjacent page objects. Despite their advancements, the performance of existing semantic segmentation-based methods remains inferior to that of the other two categories of approaches when evaluated on recent document layout analysis benchmarks.

\textbf{Graph-based methods.}
These approaches (e.g., \cite{li2018page,li2020page,luo2022doc,wang2022post}) represent each document page as a graph, where the nodes correspond to primitive page objects (e.g., words, text-lines, connected components), and the edges denote relationships between neighboring primitive page objects. The detection of page objects is then formulated as a graph labeling problem. Li et al. \cite{li2018page} employed image processing techniques to initially generate line regions, followed by the application of two CRF models to classify these regions into distinct types and predict whether pairs of line regions belong to the same instance, based on visual features extracted by CNNs. Subsequently, line regions that share the same class and instance are merged to form page objects. In their follow-up work \cite{li2020page}, Li et al. replaced line regions with connected components as nodes and implemented a graph attention network (GAT) to enhance the visual features of both nodes and edges. Luo et al. \cite{luo2022doc} concentrated on the logical role classification task, proposing the use of multi-aspect graph convolutional networks (GCNs) to identify the logical role of each page object by leveraging syntactic, semantic, density, and appearance features. More recently, Wang et al. \cite{wang2022post} focused on paragraph identification, developing a GCN-based approach to group text-lines into paragraphs. Liu et al. \cite{liu2022unified}, Long et al. \cite{long2022towards}, and Xue et al. \cite{xue2022contextual} further proposed a unified framework for text detection and paragraph (text-block) identification.

\subsection{Reading Order Prediction}
The objective of reading order prediction is to determine the appropriate reading sequence for documents. Generally, humans tend to read documents in a left-to-right and top-to-bottom manner. However, such simplistic sorting rules may prove inadequate when applied to complex documents with tokens extracted by OCR tools. Previous research has attempted to tackle the reading order issue using a variety of approaches. As categorized by Wang et al. \cite{wang2023text}, these methods can be broadly classified into rule-based sorting and machine learning-based sequence prediction, among others. 

\textbf{Rule-based sorting.}
Topological sorting, first introduced by Breuel \cite{breuel2003high}, has been utilized for document layout analysis. In this method, partial orders are determined based on the x/y interval overlaps between text lines, enabling the generation of reading order patterns for multi-column text layouts. A bidimensional relation rule, proposed in \cite{aiello2003bidimensional}, offers similar topological rules while also incorporating a row-wise rule by inverting the x/y axes from column-wise. In the same vein, an argumentation-based approach in \cite{FerilliP2015abstract} utilizes rules derived from relationships between text blocks. For text layouts with hierarchies and larger sizes, XY-Cut \cite{Meunier2005optimized, gu2022xylayoutlm} can serve as an efficient method to order all text blocks from top to bottom and left to right for specific layout types. Despite their effectiveness in certain scenarios, these rule-based methods can be prone to failure when confronted with out-of-domain cases.

\textbf{Machine learning-based sequence prediction.}
Designed to learn from training examples across various domains, machine learning-based approaches aim to provide a general solution for reading order prediction. Ceci et al. \cite{ceci2007data} introduced a probabilistic classifier within the Bayesian framework, which is capable of reconstructing single or multiple chains of layout components based on learned partial orders. Differently, an inductive logic programming (ILP) learning algorithm was applied in \cite{Malerba2007learning} to learn two kinds of predicates, \textit{first\_to\_read/1} and \textit{succ\_in\_reading/2}, thereby establishing an ordering relationship.
In recent years, deep learning models have emerged as the leading solution for numerous machine learning challenges. Li et al. \cite{li2020readingorder} proposed an end-to-end OCR text reorganizing model, using a graph convolutional encoder and a pointer network decoder to reorder text blocks. LayoutReader \cite{wang2021layoutreader} introduced a benchmark dataset called ReadingBank, which contains reading order, text, and layout information, and employed a transformer-based architecture on spatial-text features to predict the reading order sequence of words.
However, the decoding speed of these auto-regressive-based methods is limited when applied to rich text documents. Recently, Quir{'{o}}s et al. \cite{Quiros2022reading} followed the idea of assuming a pairwise partial order at the element level from \cite{breuel2003high} and proposed two new reading-order decoding algorithms for reading order prediction on handwritten documents. They also provided a theoretical background for these algorithms. A significant limitation of this approach is that the partial order between two entities is determined solely by pair-wise spatial features, without considering the visual information and textual information.

\subsection{Hierarchical Document Structure Reconstruction}
The process of reconstructing a document's hierarchical structure aims to recover its logical structure, which conveys semantic information beyond the character strings that comprise its contents. Table of Contents is a crucial component in reconstructing the hierarchical structure. Consequently, existing research studies on hierarchical structure reconstruction can be broadly categorized into two groups. The first group primarily focuses on extracting the table of contents within documents. The second group places emphasis on overall structure reconstruction of a document.

\textbf{Table of Contents.}
Table of contents extraction is the task of restoring the structure of a document and recognizing the hierarchy of its sections. It is a challenging task due to the diversity of TOC styles and layouts. Early methods relied on heuristic rules derived from small data sets for specific domains, which were not effective in large-scale heterogeneous documents. 
Wu et al. \cite{wu2013toc} identified three basic TOC styles: ``flat", ``ordered", and ``divided". Based on these styles, they proposed an approach for TOC recognition that adaptively selects appropriate rules according to the basic TOC style features. However, this method assumes the existence of a Table of Contents page within the documents. 
Nguyen et al. \cite{Nguyen2017enhancing} proposed a system that combines a TOC page detection method with a link-based TOC reconstruction method to address the TOC extraction problem. 
Cao et al. \cite{cao2022held} developed a framework called Hierarchy Extraction from Long Document (HELD) to tackle the problem of TOC extraction in long documents. This approach sequentially inserts each section heading into the TOC tree at the correct position, considering sibling and parent information using LSTM \cite{hochreiter1997long}. 
Recently, Hu et al. \cite{hu2022toc} proposed an end-to-end model by using a multimodal tree decoder (MTD) for table of contents extraction. The MTD model fuses multimodal features for each entity of the document and parses the hierarchical relationship by a tree-structured decoder. 

\textbf{Overall Structure Reconstruction.}
To reconstruct the overall structure of a document, it is critical to represent the structure and layout of the document. Intuitively, graph representation for document structure is most general and can encapsulate the relationship between regions and their properties. The graph representation, however, fails to capture the hierarchical nature of a document structure and layout. Also, it is hard to define a complete graph representation for a document. To accomplish this, one could use a rooted tree for representing document layout and logical structure \cite{nagy1984hierarchical}. One of the most powerful ways to express hierarchical structures is to use formal grammars \cite{conway1993page}. The class of regular and context-free grammars are extremely useful in describing the structure of most documents. However, there could be multiple derivations corresponding to a particular sequence of terminals. This would mean multiple interpretations of the structure or layout. Tateisi et al. \cite{tateisi1994using} proposed a stochastic grammar to integrate multiple evidences and estimate the most probable parse or interpretation of a given document. Despite its usefulness, stochastic grammars may lack the flexibility to model complex patterns and structures, particularly when handling highly diverse data. In recent years, some deep learning based methods are proposed for tree-based document structure reconstruction. Wang et al. \cite{wang2020docstruct} concentrated on form understanding task, treating the form structure as a tree-like hierarchy composed of text fragments. To predict the relationship between each pair of text fragments, they employed an asymmetric parameter matrix. However, this approach resulted in high computational complexity when dealing with documents containing a large number of text fragments. DocParser, as proposed by Rausch et al. \cite{rausch2021docparser}, presented an end-to-end system designed to parse the complete physical structure of documents including all text elements, nested figures, tables, and table cell structures. This system employed rule-based algorithms for relation classification and inferred document structures in a holistic, principled manner. Nonetheless, the system did not consider the logical hierarchical structure of documents, such as the table of contents, and the reliance on a rule-based approach significantly constrained its overall effectiveness and adaptability. Recently, Ma et al. \cite{Ma2023HRDoc} introduced hierarchical reconstruction of document structures as a novel task and built a large-scale dataset, named HRDoc. Moreover, an encoder-decoder-based hierarchical document structure parsing system (DSPS) was proposed to reconstruct the hierarchical structure. While taking into account the logical structure of the document, this task presumes that the reading order is provided. Furthermore, DSPS directly predicts relationships between text-lines, resulting in low representational ability and high computational cost. In this work, we also consider overall structure reconstruction to be a recovery of the structure of the rooted tree of a document.
\section{Problem Definition}

\begin{figure}[t]
    \centering
    \includegraphics[width=0.9\linewidth]{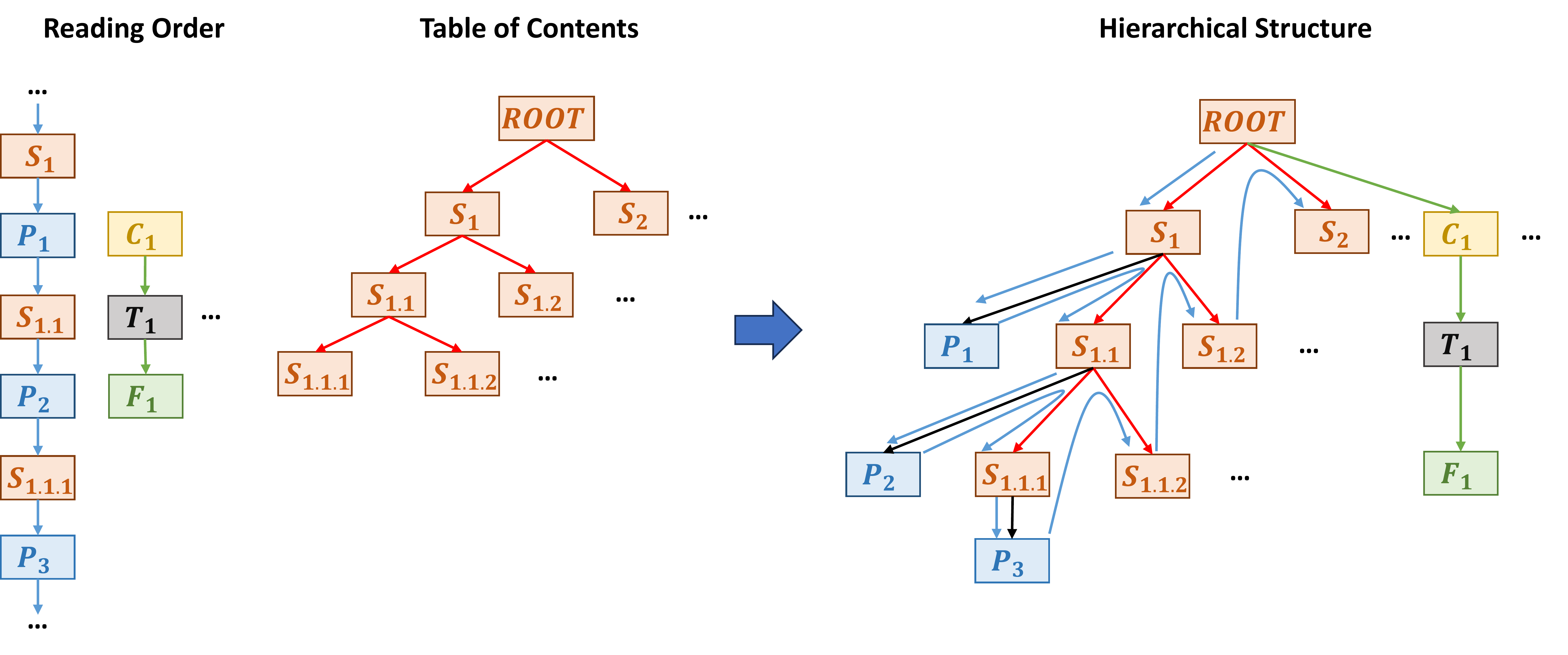}
    \caption{Hierarchical structure reconstruction of a document by integrating the Reading Order and Table of Contents. Blue arrows demonstrate the Text Region Reading Order Relationship, green arrows show the Graphical Region Relationship, and red arrows signify the TOC Relationship. The nodes ``P", ``S", ``C", ``T" and ``F" represent Paragraph, Section heading, Caption, Table and Footnote, respectively.}
    \label{fig:reconstruct}
\end{figure}

The majority of document types, such as scientific papers, books, reports, and legal documents, typically exhibit a hierarchical document structure in a tree-like format. In this structure, the nodes within the tree represent various page objects (e.g., section, paragraph, figure, caption) of the document, while the edges signify the hierarchical relationships and connections between these page objects. Given a multi-page document $D$ comprised of $D_1, D_2, ..., D_n$, where $D_i$ represents an individual page within document $D$, the primary objective of hierarchical document structure analysis is to reconstruct its hierarchical structure tree $H$, consisting of both page objects and hierarchical relationships as follows:

\textbf{Page Objects} ($O_i, i=1,...,m$) refer to the various page objects within document $D$. Each page object is described by three attributes: 1) its logical role category $\mathbf{c}_i \in C $ (e.g., title, section heading, table, figure, etc.); 2) its bounding box coordinates $\mathbf{b}_i$; 3) its basic semantic units (not useful for graphical page objects and we use OCR'd text-lines as basic semantic units). 

\textbf{Hierarchical Relationships} ($R_{ij}, i,j=1,...,m$) describe the relationships between page object pairs and are represented by triplets $(O_{i}, \boldsymbol{r}_{ij}, O_{j})$. Each triplet includes a subject page object $O_{i}$, an object page object $O_{j}$, and a relation type $\boldsymbol{r}_{ij} \in \Phi$.  Based on the categories of $O_i$ and $O_j$, we define the following three relationship types: 1) \textbf{Text Region Reading Order Relationship} between main body text regions, 2) \textbf{Graphical Region Relationship} between caption, footnote and graphical page objects, i.e., tables or figures; 3) \textbf{Table of Contents Relationship} between section heading regions.

The combination of page objects and hierarchical relationships is sufficient to reconstruct the hierarchical tree $H$ for a document, as illustrated in Fig.~\ref{fig:reconstruct}. Conversely, the hierarchy tree $H$ can be used to extract various hierarchical relationships as needed, further emphasizing its importance in the process of hierarchical document structure analysis. For instance, the reading order sequence can be obtained by performing a pre-order traversal on the hierarchical tree $H$. Based on the problem description and objectives of hierarchical document structure analysis, we divide it into following three distinct sub-tasks, which correspond to our proposed three-stage framework:

\begin{itemize}
    \item \textbf{Page Object Detection} (Detect stage) aims to identify individual page object $O_i$ (e.g., text regions, images, tables) within each page of the document $D$ and assign a logical role to each detected page object (e.g., section headings, captions, footnotes).
    \item \textbf{Reading Order Prediction} (Order stage) focuses on determining the reading sequence of detected page objects based on their spatial arrangement within the document $D$. The reading order is represented as a permutation of the indices of the detected page objects.
    \item \textbf{Table of Contents Extraction} (Construct stage) aims to extract the table of contents within document $D$, which involves constructing a hierarchy tree that summarizes the overall hierarchical structure $H$. The hierarchy tree comprises a list of section headings and their hierarchical levels.
\end{itemize}

By integrating the results from all three sub-tasks, the hierarchical document structure tree $H$ can be effectively reconstructed, offering a more comprehensive understanding of complex documents.

\section{Methodology}

\subsection{Overview}
Our newly proposed tree construction based approach for hierarchical document structure analysis, named Detect-Order-Construct, is illustrated in Fig.~\ref{fig:pipeline}. This approach comprises three main components: 1) A \textbf{Detect} stage that identifies individual page objects within the document rendering and assigns a logical role to each detected page object (i.e., page object detection); 2) An \textbf{Order} stage responsible for determining the sequential order of the page objects (i.e., reading order prediction); and 3) A \textbf{Construct} stage that extracts the abstract hierarchy tree (i.e., table of contents extraction). By integrating the outputs from the previous tasks, we can effectively reconstruct a complete hierarchical document structure tree (i.e., hierarchical document structure reconstruction).

In our approach, we uniformly define the tasks of these three stages as relation prediction problems and present a type of multi-modal, transformer-based relation prediction models to tackle all tasks effectively. Our proposed relation prediction model approaches relation prediction as a dependency parsing task and incorporates structure-aware designs that align with the chain structure of reading order and the tree structure of table of contents. Utilizing our novel techniques and the proposed framework, we develop an effective end-to-end solution for hierarchical document structure analysis, which comprises three modules: the Detect module, the Order module, and the Construct module. We elaborate on the details of these three modules in Sections \ref{subsec:group}, \ref{subsec:order}, and \ref{subsec:reconstruct}, respectively.

\subsection{Detect Module}
\label{subsec:group}
\begin{figure}[t]
    \centering
    \includegraphics[width=1.0\textwidth]{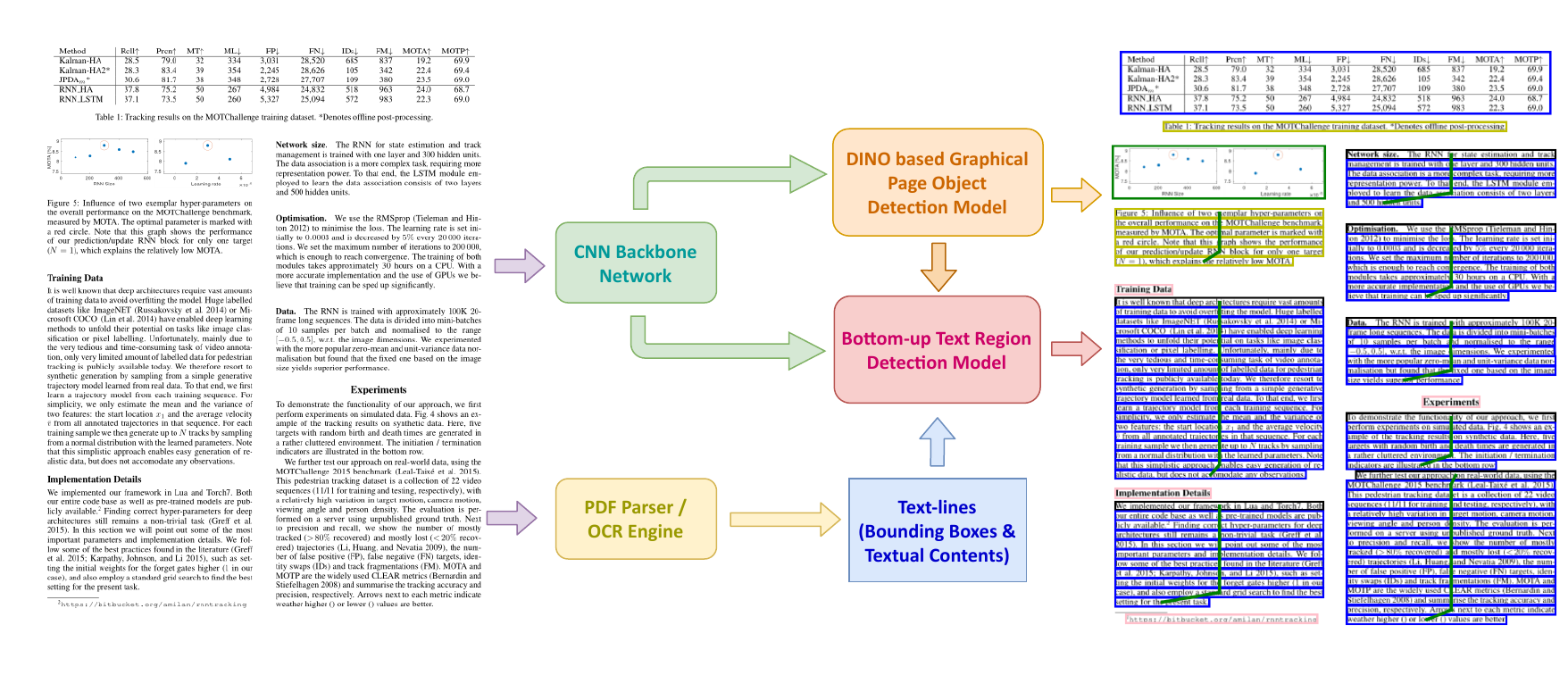}
    \caption{The overall architecture of our Detect module.}
    \label{fig:group}
\end{figure}

The proposed Detect module consists of three primary components: 1) A shared visual backbone network designed to extract multi-scale feature maps from input document images; 2) A top-down graphical page object detection model for detecting graphical page objects, such as tables, figures, and displayed formulas; 3) A bottom-up text region detection model that groups text-lines located outside graphical page objects into text regions, based on the intra-region reading order, and identifies the logical role of each text region. The overall architecture of the Detect module is illustrated in Fig.~\ref{fig:group}. 
In our conference paper \cite{zhong2023hybrid}, we selected a ResNet-50 network as the backbone network to generate multi-scale feature maps and the DINO \cite{zhang2023dino} as the top-down graphical page object detector to localize these graphical objects. However, any suitable visual backbone network and object detection or instance segmentation model can be readily incorporated into our Detect module. In this paper, we primarily concentrate on the details of the newly proposed \textit{Bottom-up Text Region Detection Model}.

\begin{figure}[t]
    \centering
    \includegraphics[width=1.0\linewidth]{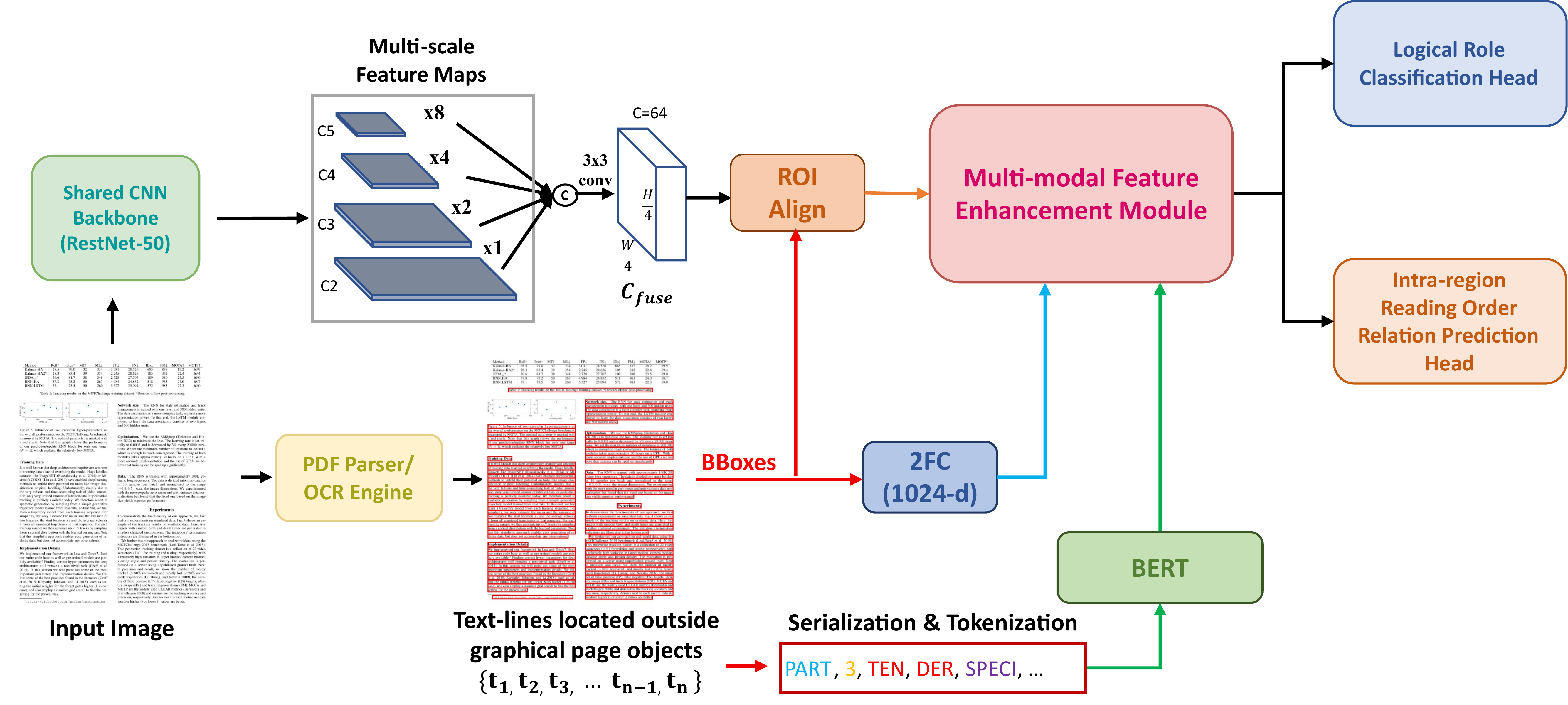}
    \caption{A schematic view of the proposed bottom-up text region detection model.}
    \label{fig:textline-grouping}
\end{figure}

\begin{figure}[t]
    \centering
    \includegraphics[width=1.0\linewidth]{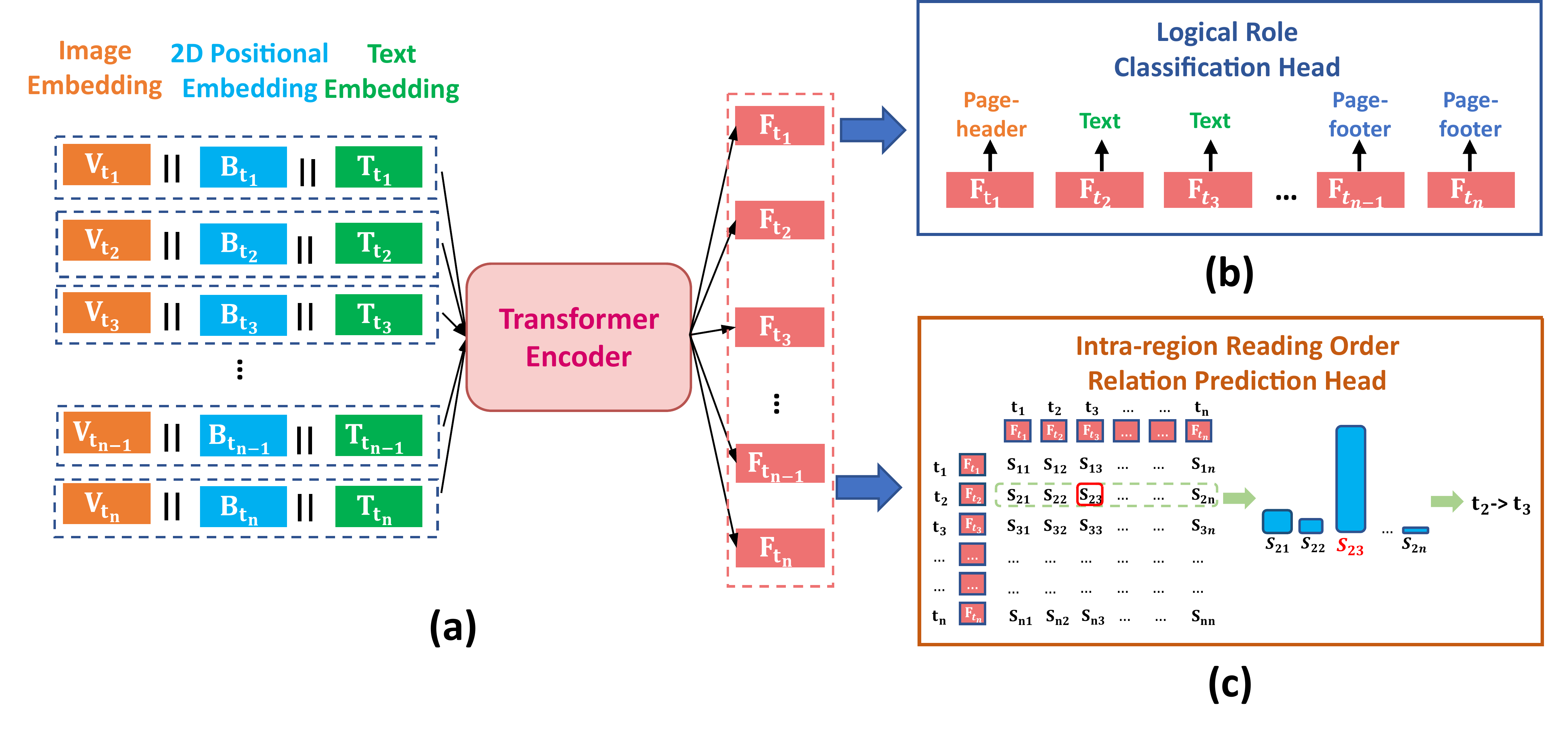}
    \caption{Illustration of (a) Multi-modal Feature Enhancement Module; (b) Logical Role Classification Head; (c) Reading Order Relation Prediction Head in bottom-up text region detection model.}
    \label{fig:textline-grouping-part2}
\end{figure}

A text region is a semantic unit of writing that comprises a group of text-lines arranged in natural reading order and associated with a logical label, such as paragraph, list/list-item, title, section heading, header, footer, footnote, and caption. Given a document page rendering $D_i$ composed of $n$ text-lines $[t_1, t_2, ..., t_n]$, the objective of our bottom-up text region detection model is to group these text-lines into distinct text regions according to the intra-region reading order and to recognize the logical role of each text region. In this study, we assume that the bounding boxes and textual contents of text-lines have already been provided by a PDF parser or OCR engine. Based on the detection results of the top-down graphical page object detection model, we initially filter out those text-lines located inside graphical page objects and then utilize the remaining text-lines as input. As depicted in Fig.~\ref{fig:textline-grouping}, our bottom-up text region detection model consists of a multi-modal feature extraction module, a multi-modal feature enhancement module, and two prediction heads, i.e., an intra-region reading order relation prediction head and a logical role classification head. The detailed illustrations of the multi-modal feature enhancement module and the two prediction heads can be found in Fig. \ref{fig:textline-grouping-part2}.

\subsubsection{Multi-modal Feature Extraction Module} 
\label{sec:feat_extract}
In this module, we extract the visual embedding, text embedding, and 2D Positional Embedding for each text-line. 

\textbf{Visual Embedding.} As shown in Fig.~\ref{fig:textline-grouping}, we first resize $C_4$ and $C_5$ to the size of $C_3$ and then concatenate these three feature maps along the channel axis, which are fed into a $3\times3$ convolutional layer to generate a feature map $C_{fuse}$ with 256 channels.
For each text-line $t_i$, we adopt the RoIAlign algorithm \cite{he2017mask} to extract $7\times7$ feature maps from $C_{fuse}$ based on its bounding box $b_{t_i} = (x_i^1,y_i^1,x_i^2,y_i^2)$, where $(x_i^1,y_i^1)$, $(x_i^2,y_i^2)$ represent the coordinates of its upper left and bottom right corners, respectively. The final visual embedding $V_{t_i}$ of $t_i$ can be represented as:
\begin{equation}
    V_{t_i} = LN(ReLU(FC(ROIAlign(C_{fuse}, b_{t_i})))),
    \label{equ:visual}
\end{equation}
where FC is a fully-connected layer with 1,024 nodes and LN represents Layer Normalization \cite{ba2016layer}.

\textbf{Text Embedding.} We leverage the pre-trained language model BERT \cite{devlin2018bert} to extract the text embedding of each text-line. Specifically, we first serialize all the text-lines in a document image into a 1D sequence by reading them in a top-left to bottom-right order and tokenize the text-line sequence into a sub-word token sequence, which is then fed into BERT to get the embedding of each token. After that, we average the embeddings of all the tokens in each text-line $t_i$ to obtain its text embedding $T_{t_i}$, followed by a fully-connected layer with 1,024 nodes to make the dimension the same as that of $V_{t_i}$:
\begin{equation}
\label{equ:text}
    T_{t_i} = LN(ReLU(FC(T_{t_i})))).
\end{equation}

\textbf{2D Positional Embedding.} For each text-line $t_i$, we encode its bounding box and size information as its 2D Positional Embedding $B_{t_i}$:
\begin{equation}
    B_{t_i} = LN(MLP(x_i^1 / W, y_i^1 / H, x_i^2 / W, y_i^2 / H, w_i / W, h_i/ H)),
    \label{equ:pos}
\end{equation}
where $(w_i,h_i)$ and $(W,H)$ represent the width and height of $b_{t_i}$ and the input image, respectively. MLP consists of 2 fully-connected layers with 1,024 nodes, each of which is followed by ReLU.

For each text-line $t_i$, we concatenate its visual embedding $V_{t_i}$, text embeddings $T_{t_i}$, and 2D Positional Embedding $B_{t_i}$ to obtain its multi-modal representation $U_{t_i}$.
\begin{equation}
    U_{t_i} = FC(Concat(V_{t_i}, T_{t_i}, B_{t_i})), 
\end{equation}
where FC is a fully-connected layer with 1,024 nodes.

\subsubsection{Multi-modal Feature Enhancement Module} 
\label{group:enhance}
As shown in Fig.~\ref{fig:textline-grouping-part2}, we use a lightweight Transformer encoder to further enhance the multi-modal representations of text-lines by modeling their interactions with a self-attention mechanism. Each text-line is treated as a token of the Transformer encoder and its multi-modal representation is taken as the input embedding:
\begin{equation}
    F_t = TransformerEncoder(U_t),
    \label{equ:textlines_feat}
\end{equation}
where $U_t=[U_{t_1}, U_{t_2},..., U_{t_n}]$ and $F_t=[F_{t_1}, F_{t_2},..., F_{t_n}]$ are the input and output embeddings of the Transformer encoder, $n$ is the number of the input text-lines. To save computation, here we only use a 1-layer Transformer encoder, where the head number, dimension of hidden state, and the dimension of feedforward network are set as 12, 768, and 2048, respectively.

\subsubsection{Intra-region Reading Order Relation Prediction Head}
\label{sec:reading}
We propose to use a relation prediction head to predict intra-region reading order relationships between text-lines. Given a text-line $t_i$, if a text-line $t_j$ is its succeeding text-line in the same text region, we define that there exists an intra-region reading order relationship ($t_i \rightarrow t_j$) pointing from text-line $t_i$ to text-line $t_j$. If text-line $t_i$ is the last (or only) text-line in a text region, its succeeding text-line is considered to be itself. 
Unlike many previous methods that consider relation prediction as a binary classification task \cite{li2018page,wang2022post}, 
we treat relation prediction as a dependency parsing task and use a softmax cross-entropy loss to replace the standard binary cross-entropy loss during optimization by following \cite{zhang2021entity}. Moreover, we adopt a spatial compatibility feature introduced in \cite{zhang2017relationship} to effectively model spatial interactions between text-lines for relation prediction.

Specifically, we use a multi-class (i.e., $n$-class) classifier to calculate a score $s_{ij}$ to estimate how likely $t_j$ is the succeeding text-line of $t_i$ as follows:
\begin{gather}
\label{relation_score}
    f_{ij} = FC_q(F_{t_i}) \circ FC_k(F_{t_j}) + MLP(r_{b_{t_i},b_{t_j}}), \\
    s_{ij} = \frac{\exp(f_{ij})}{\sum_N \exp(f_{ij})},
\end{gather}
where each of $FC_q$ and $FC_k$ is a single fully-connected layer with 2,048 nodes to map $F_{t_i}$ and $F_{t_j}$ into different feature spaces; $\circ$ denotes dot product operation; MLP consists of 2 fully-connected layers with 1,024 nodes and 1 node respectively;
$r_{b_{t_i},b_{t_j}}$ is a spatial compatibility feature vector between ${b_{t_i}}$ and ${b_{t_j}}$, which is a concatenation of three 6-d vectors:
\begin{equation}
\label{rel2d_1}
r_{b_{t_i},b_{t_j}} = (\Delta(b_{t_i},b_{t_j}), \Delta(b_{t_i}, b_{t_{ij}}), \Delta(b_{t_j}, b_{t_{ij}})),
\end{equation}
where $b_{t_{ij}}$ is the union bounding box of $b_{t_i}$ and $b_{t_j}$; $\Delta(.,.)$ represents the box delta between any two bounding boxes. Taking $\Delta(b_{t_i}, b_{t_j})$ as an example, $\Delta(b_{t_i}, b_{t_j}) = ( d^{x_{\text{ctr}}}_{ij}, d^{y_{\text{ctr}}}_{ij}, d^w_{ij}, d^h_{ij}, d^{x_{\text{ctr}}}_{ji}, d^{y_{\text{ctr}}}_{ji})$, where each dimension is given by:
\begin{equation}
\begin{aligned}
     d^{x_{\text{ctr}}}_{ij} &= (x^{\text{ctr}}_{i} - x^{\text{ctr}}_{j})/w_{i}, &d^{y_{\text{ctr}}}_{ij} &=  (y^{\text{ctr}}_{_i} - y^{\text{ctr}}_{j})/h_{i}, \\
     d^w_{ij} &= \log (w_{i}/w_{j}), &d^h_{ij} &= \log (h_{i}/h_{j}), \\
     d^{x_{\text{ctr}}}_{ji} &= (x^{\text{ctr}}_{j} - x^{\text{ctr}}_{i})/w_{j}, &d^{y_{\text{ctr}}}_{ji} &=  (y^{\text{ctr}}_{j} - y^{\text{ctr}}_{i})/h_{j},
\end{aligned}
\end{equation}
where $(x^{\text{ctr}}_{i}, y^{\text{ctr}}_{i})$ and $(x^{\text{ctr}}_{j}, y^{\text{ctr}}_{j})$ are the center coordinates of $b_{t_i}$ and $b_{t_j}$, respectively.

We select the highest score from scores $[s_{ij}, j=1,2,...,n]$ and output the corresponding text-line as the succeeding text-line of $t_i$. To achieve higher relation prediction accuracy for the intra-region reading order relationship, which has a chain structure, we employ an additional relation prediction head to further identify the preceding text-line for each text-line. The prediction results from both relation prediction heads are then combined to obtain the final results. Based on the predicted intra-region reading order relationships, we group text-lines into text regions using a Union-Find algorithm. The bounding box of the text region is the union bounding box of all its constituent text-lines.

\subsubsection{Logical Role Classification Head}

Given the enhanced multi-modal representations of text-lines $F_t=[F_{t_1}, F_{t_2},..., F_{t_n}]$, we add a multi-class classifier to predict a logical role label for each text-line and determine the logical role of each text region by the plurality voting of all its constituent text-lines.

\subsection{Order Module}
\label{subsec:order}

\begin{figure}[t]
    \centering
    \includegraphics[width=1.0\linewidth]{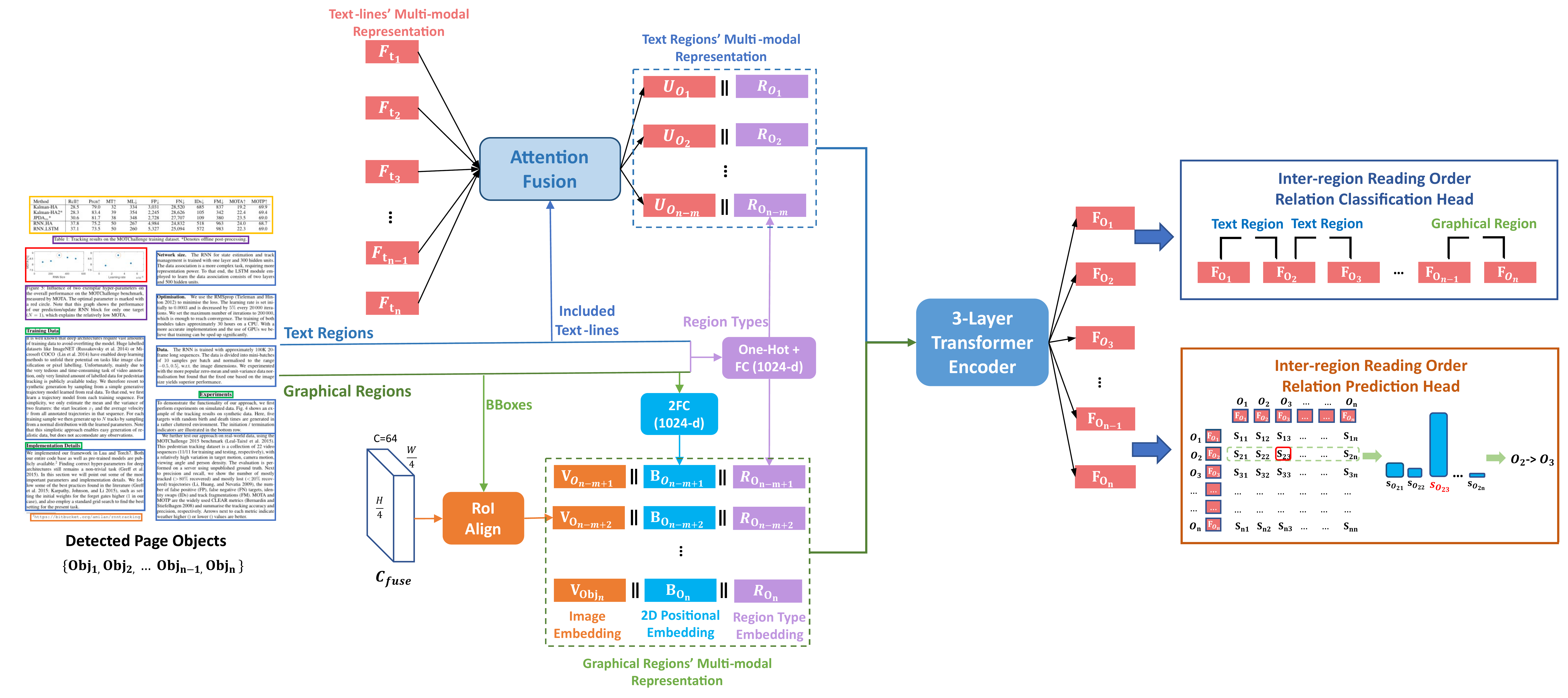}
    \caption{Architecture of our proposed Order module for reading order prediction.}
    \label{fig:order}
\end{figure}

The Order module focuses on determining the reading sequence of graphical page objects and text regions identified by the Detect module within document $D$. Similar to the bottom-up text region detection model employed in the Detect module, we also utilize our proposed multi-modal, transformer-based relation prediction model to predict the inter-region reading order relationships among the recognized page objects. The Order module processes the detected page objects as input and employs an attention-based approach to integrate the features of text-lines belonging to the same text region, thereby achieving a more efficient feature representation of the text region. Furthermore, we define two categories of inter-region reading order relationships: (1) Text region reading order relationships between main body text regions, (2) Graphical region reading order relationships between captions/footnotes and graphical page objects such as tables and figures. Consequently, we incorporate an additional inter-region reading order relation classification head to predict relation types. A detailed illustration of the Order module can be found in Fig.~\ref{fig:order}.

\subsubsection{Multi-modal Feature Extraction Module} 

Following Eqs. (\ref{equ:visual}) and (\ref{equ:pos}) as described in Section \ref{sec:feat_extract}, we fuse the visual embedding and the 2D positional embedding to obtain a multi-modal representation $U_{O_{m}}$ for each graphical page object $O_m$ in a similar manner. For each detected text region page object $O_n$ consisting of text-lines $[t_{n_1}, t_{n_2}, ..., t_{n_k}]$, we propose an attention fusion model to integrate the features of text-lines $[F_{t_{n_1}}, F_{t_{n_2}}, ..., F_{t_{n_k}}]$ produced by Eq. (\ref{equ:textlines_feat}), thereby forming a multi-modal representation $U_{O_{n}}$ for this text region as follows:
\begin{gather}
    \alpha_{t_{n_j}} = FC_1(tanh(FC_2(F_{t_{n_j}}))), \\
    w_{t_{n_j}} = \frac{\exp \alpha_{t_{n_j}}}{\sum_{j} \exp \alpha_{t_{n_j}}}, \\
    U_{O_{n}} = \sum_{j} w_{t_{n_j}} F_{t_{n_j}},
\end{gather}
where both $FC_1$ and $FC_2$ are single fully-connected layers with 1,024 and 1 nodes, respectively. Furthermore, for each page object, we derive a region type embedding for each page object as follows:
\begin{equation}
    R_{O_i} = LN(ReLU(FC(Embedding(r_{O_i})))),
\end{equation}
where $Embedding$ is an embedding layer with 1,024 hidden dimension and $r_{O_i}$ is the logical role of the page object $O_i$.

Lastly, we concatenate each page object's multi-modal representation $U_{O_{i}}$ and region type embedding $R_{O_i}$ to obtain its final representation $\hat{U}_{O_i}$ as follows:
\begin{equation}
    \hat{U}_{O_i} = FC(Concat(U_{O_{i}}, R_{O_i})), 
\end{equation}
where $FC$ is a fully-connected layer with 1,024 nodes.

\subsubsection{Multi-modal Feature Enhancement Module} 
\label{order:enhance}
As illustrated in Fig.~\ref{fig:order}, we adopt a similar approach to previous multi-modal feature enhancement module in the Group stage. In this case, we utilize a three-layer Transformer encoder to further improve the multi-modal representations of page objects by modeling their interactions using a self-attention mechanism. Each page object is treated as a token of the Transformer encoder, and its multi-modal representation serves as the input embedding:
\begin{equation}
    F_{O} = TransformerEncoder(\hat{U}_{O}),
\end{equation}
where $\hat{U}_{O}=[\hat{U}_{O_i}, \hat{U}_{O_2},..., \hat{U}_{O_n}]$ and $F_{O}=[F_{O_1}, F_{O_2},..., F_{O_n}]$ represent the input and output embeddings of the Transformer encoder, and $n$ is the number of the input page objects. The hyperparameters of the transformer encoder are consistent with those in the Detect module, except for the layer number.

\subsubsection{Inter-region Reading Order Relation Prediction Head}

Owing to the similarity between the inter-region reading order task of the Order module and the intra-region reading order task of the Detect module, we employ an identical structure for the inter-region reading order relation prediction head in both modules. Further details about this head can be found in Section \ref{sec:reading}.

\subsubsection{Inter-region Reading Order Relation Classification Head}

We employ a multi-class classifier to compute the probability distribution across various classes in order to determine the relation type between page object $O_i$ and page object $O_j$. It works as follows:
\begin{gather}
p_{ij} = BiLinear(FC_q(F_{O_i}), FC_k(F_{O_j})), \\
c_{ij} = argmax(p_{ij}),
\end{gather}
where both $FC_q$ and $FC_k$ represent single fully-connected layers with 2,048 nodes, which are used to map $F_{O_i}$ and $F_{O_j}$ into distinct feature spaces; $BiLinear$ signifies the bilinear classifier; and $argmax$ refers to identifying the index $c_{ij}$ of the maximum value within the given probability distribution $p_{ij}$ as the predicted relation type.

\subsection{Construct Module}
\label{subsec:reconstruct}

\begin{figure}[t]
    \centering
    \includegraphics[width=1.0\linewidth]{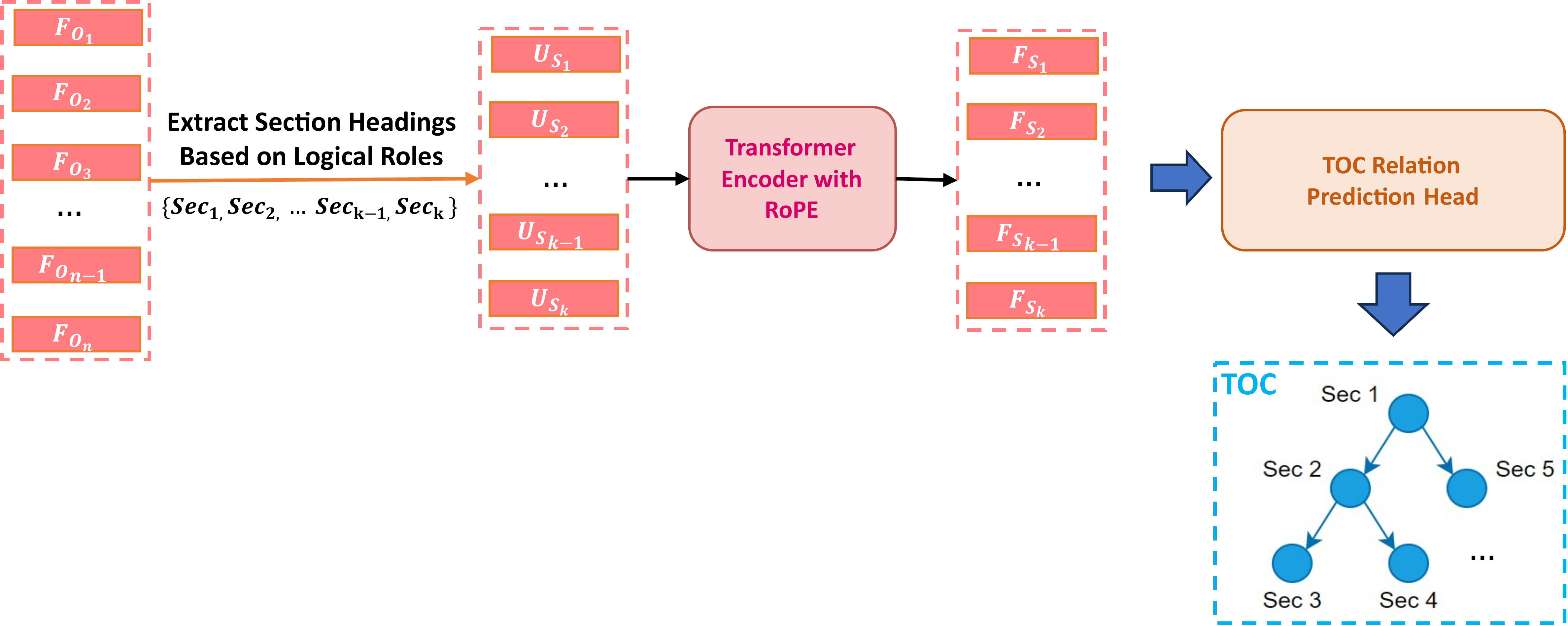}
    \caption{Illustration of the Construct module.}
    \label{fig:toc}
\end{figure}

\begin{figure}[t]
    \centering
    \includegraphics[width=1.0\linewidth]{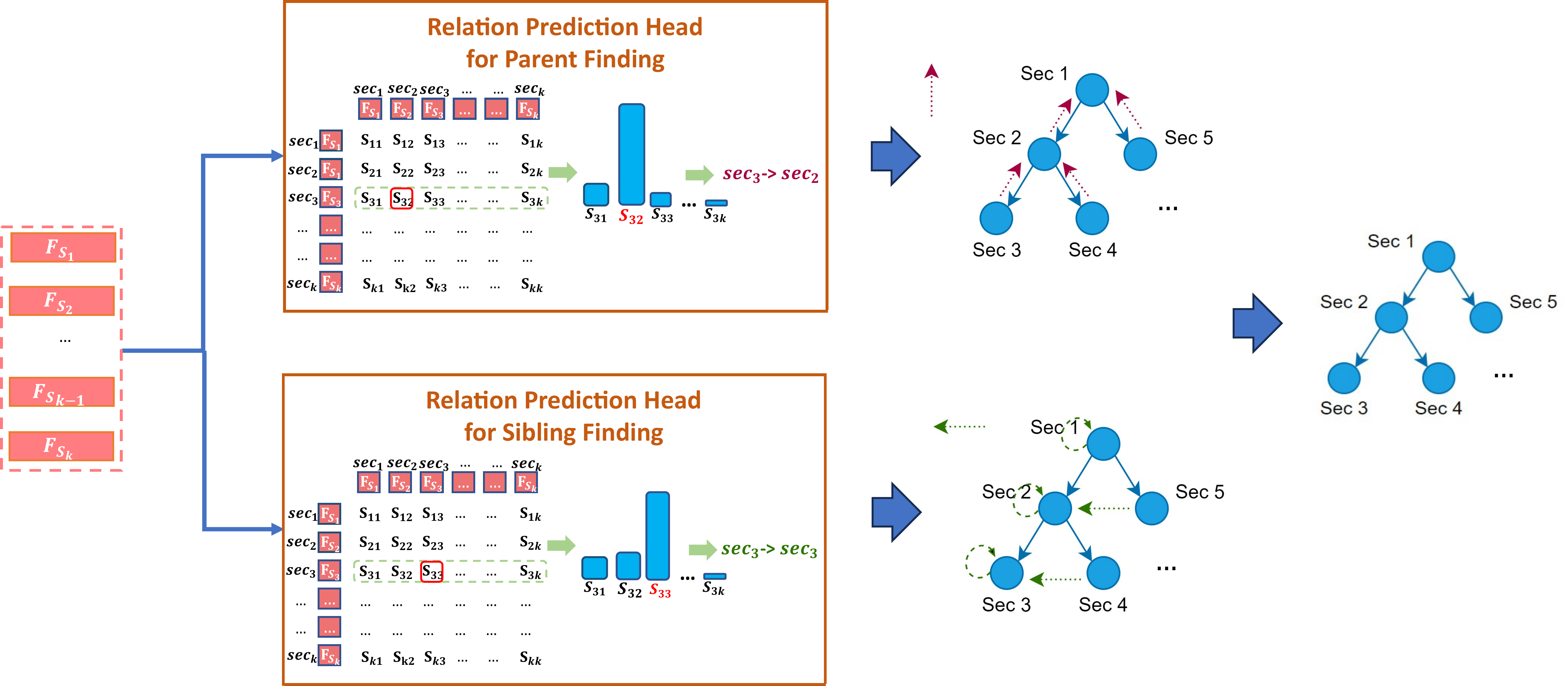}
    \caption{Illustration of TOC Relation Prediction Head.}
    \label{fig:toc_head}
\end{figure}

Given the detected section headings $[sec_1, sec_2, ..., sec_{k-1}, sec_k]$ arranged according to the predicted reading order sequence for document $D$, the goal of the Construct module is to generate a tree structure representing the hierarchical table of contents. As illustrated in Fig.~\ref{fig:toc}, we extract the multi-modal representation $F_{S_i}$ of each section heading $sec_i$ from all page objects' multi-modal representation $F_{O}$ based on the logical role. Subsequently, we input all section headings' representation $U_{S} = [U_{S_1}, U_{S_2}, ..., U_{S_k}]$ into a transformer encoder to further enhance the representations. However, unlike the transformer encoder employed in the Detect module and the Order module, both of which are order-agnostic, the input sequence $U_{S}$ has the correct reading order predicted by the Order module, allowing us to add a positional encoding to convey the reading order information. To incorporate the relative position in the reading order sequence and accommodate a larger scale of page numbers in the document, we utilize an efficient positional encoding method called Rotary Positional Embedding (RoPE) \cite{su2021roformer}. RoPE encodes the absolute position using a rotation matrix and simultaneously includes the explicit relative position dependency in the self-attention formulation. Following the Multi-modal Feature Enhancement Module, we generate the enhanced representations $F_{S} = [F_{S_1}, F_{S_2}, ..., F_{S_k}]$ for section headings. Finally, we introduce a tree-aware TOC relation prediction head to predict the TOC relationships among these section headings. The specially designed relation prediction head is illustrated in Fig.~\ref{fig:toc_head}.

\subsubsection{TOC Relation Prediction Head}

During the generation of the ordered tree for Table of Contents, solely relying on the relationship features between child and parent nodes has proven to be insufficient. Some prior studies \cite{hu2022toc, Ma2023HRDoc, cao2022held} have already observed that incorporating information from sibling nodes can lead to an improved generation of the TOC. Inspired by these works, we propose two types of TOC relationships between section heads to further enhance the TOC generation process: parent-child relationships and sibling relationships.

The parent-child relationship is relatively straightforward: when a section heading $sec_i$ serves as the parent node for another section heading $sec_j$ within the TOC tree structure, we define a parent-child relationship $(sec_j\rightarrow sec_i)$ that points from $sec_j$ to $sec_i$. Sibling relationships in a TOC tree are established as follows: if section heading $sec_i$ acts as the left sibling of section heading $sec_j$, then a sibling relationship $(sec_j\rightarrow sec_i)$ is present. In cases where a section heading lacks a parent node or left sibling node, its parent-child or sibling relationship is defined as pointing to itself. This approach aims to provide a more comprehensive representation of the relationships among section heads, ultimately leading to a more accurate and robust TOC generation.

As illustrated in Fig.~\ref{fig:toc_head}, our proposed TOC Relation Prediction Head comprises two distinct relation prediction heads for the parent-child and sibling relationships, respectively. Both relation prediction heads in our proposed module employ the same network structure. To elaborate, we use the relation prediction head for the parent-child relationship as an example. Specifically, we implement a multi-class (k-class) classifier to compute a score $s^{p}_{ij}$, which estimates the likelihood of $sec_j$ being the parent node of $sec_i$. The calculation is as follows:
\begin{gather}
\label{relation_score}
f_{ij} = FC_q(F_{S_i}) \circ FC_k(F_{S_j}), \\
s^{p}_{ij} = \frac{\exp(f_{ij})}{\sum_j \exp(f_{ij})},
\end{gather}
where each of $FC_q$ and $FC_k$ represents a single fully-connected layer with 2,048 nodes to map $F_{S_i}$ and $F_{S_j}$ into distinct feature spaces; $\circ$ denotes the dot product operation. Similarly, we can obtain the score $s^{s}_{ij}$ to estimate the likelihood of $sec_j$ being the defined sibling node of $sec_i$. This unified network structure allows for efficient and effective prediction of relationships between section heads, contributing to the overall TOC generation process.

In a manner similar to the previously proposed reading order relation prediction head in Section \ref{sec:reading}, we treat relation prediction as a dependency parsing task and employ a softmax cross-entropy loss instead of the standard binary cross-entropy loss during the training phase. During the testing phase, we utilize serial decoding to integrate the outputs of the two relation prediction heads and introduce a tree structure constraint to enhance the final prediction results. Specifically, assuming that $[sec_1, sec_2, ..., sec_k]$ has been sorted according to the predicted reading order, we initialize a tree T containing only one root node, $ROOT$. Subsequently, we devise a tree insertion algorithm, as detailed in Algorithm \ref{alg:insert}, to insert each section heading in order, ultimately generating a complete table of contents tree. This approach ensures that the predicted relationships between section headings are consistent with the hierarchical tree structure, resulting in a more accurate and coherent TOC.

\begin{algorithm}
\caption{Tree Insertion Algorithm}
\label{alg:insert}
\begin{algorithmic}[1]
\REQUIRE Empty Tree $T=\{ROOT\}$, Ordered Section Headings $[sec_1, sec_2, ..., sec_k]$, 
\\ \quad \quad \quad Parent Score Matrix $\mathbf{s^{p}}$, Sibling Score Matrix $\mathbf{s^{s}}$
\FOR{$i = 1$ \TO $k$}
\STATE Find the rightmost sub-tree of $T$ and retrieve nodes $[sec_{r_1}, sec_{r_2}, \dots, sec_{r_n}]$
\STATE Compute the parent score vector $\mathbf{scores_p} = \mathbf{s^{p}}[sec_i, [sec_{r_1}, sec_{r_2}, \dots, sec_{r_n}]] \in \mathbb{R}^n$
\STATE Compute the sibling score vector $\mathbf{scores_s} = \mathbf{s^{s}}[sec_i, [sec_{r_2}, \dots, sec_{r_n}, sec_{i}]] \in \mathbb{R}^n$
\STATE Compute the final score vector $\mathbf{scores} = \mathbf{scores_p} \circ \mathbf{scores_s} \in \mathbb{R}^n$
\STATE Find the index $m$ corresponding to the maximum score in $\mathbf{scores}$
\STATE Insert $sec_i$ as the right-most child of $sec_{r_m}$
\STATE Update Tree $T$
\ENDFOR
\RETURN TOC Tree $T$
\end{algorithmic}
\end{algorithm}

\section{Experiments}

\subsection{Datasets and Evaluation Protocols}

In our conference paper \cite{zhong2023hybrid}, we conducted experiments on two widely-recognized large-scale document layout analysis benchmarks, namely PubLayNet \cite{zhong2019publaynet} and DocLayNet \cite{pfitzmann2022doclaynet} to validate the effectiveness of our proposed Detect module. In this paper, we carry out extensive experiments on a high-quality public hierarchical document structure reconstruction benchmark, HRDoc \cite{Ma2023HRDoc}, to validate the effectiveness of our proposed tree construction based framework. It is important to note that HRDoc solely provides annotations and benchmarks for the logical role classification task and the overall hierarchical structure reconstruction task. However, each sub-task plays a crucial role in hierarchical document structure analysis. Consequently, during the performance evaluation phase, conducting a thorough and rigorous assessment of each involved sub-task is essential. 

To address this issue, we expand upon the foundation of HRDoc and develop a comprehensive benchmark called Comp-HRDoc for hierarchical document structure analysis, which simultaneously evaluates page object detection, reading order prediction, table of contents extraction, and hierarchical document structure reconstruction. It is worth noting that the logical role classification in HRDoc is actually text-line-level, which may not be a fair performance evaluation for top-down approaches. Therefore, we replace it with a more popular and significant subtask, termed page object detection, in our proposed benchmark. To the best of our knowledge, Comp-HRDoc is the first benchmark designed to assess such a diverse array of document structure analysis subtasks. Our proposed model has been rigorously evaluated on this benchmark, further demonstrating the superiority of our approach.

\textbf{PubLayNet} \cite{zhong2019publaynet} is a large-scale dataset for document layout analysis released by IBM that contains 340,391, 11,858, and 11,983 document pages for training, validation, and testing, respectively. All the documents in this dataset are scientific papers publicly available on PubMed Central, and all the ground-truths are automatically generated by matching the XML representations and the content of the corresponding PDF files. It predefines 5 types of page objects, including Text (i.e., Paragraph), Title, List, Figure, and Table. The evaluation metric for PubLayNet is the COCO-style mean average precision (mAP) at multiple intersection over union (IoU) thresholds between 0.50 and 0.95 with a step of 0.05. 

\textbf{DocLayNet} \cite{pfitzmann2022doclaynet} is a challenging human-annotated document layout analysis dataset newly released by IBM that contains 69,375, 6,489, and 4,999 document pages for training, testing, and validation, respectively. It covers a variety of document categories, including financial reports, patents, manuals, laws, tenders, and scientific papers. It predefines 11 types of page objects, including Caption, Footnote, Formula, List-item, Page-footer, Page-header, Picture, Section-header, Table, Text (i.e., Paragraph), and Title. The evaluation metric for DocLayNet is also the COCO-style mean average precision (mAP), consistent with that of PubLayNet.

\textbf{HRDoc} \cite{Ma2023HRDoc} is a human-annotated dataset specifically designed to facilitate hierarchical document structure reconstruction. It features line-level annotations and cross-page relations, aiming to recover the semantic structure of PDF documents. In order to accommodate various layout types, the HRDoc dataset is divided into two parts. The first part, \textbf{HRDoc-Simple} (HRDS), consists of 1,000 documents exhibiting similar layouts. The second part, \textbf{HRDoc-Hard} (HRDH), encompasses 1,500 documents with diverse layouts. This heterogeneous collection of documents offers researchers an extensive resource to develop and assess algorithms for hierarchical document structure reconstruction in PDF documents. 

Two evaluation tasks are associated with HRDoc, including semantic unit classification (i.e., logical role classification) and hierarchical structure reconstruction. For the semantic unit classification task, the F1 score for each logical role serves as the evaluation metric. Meanwhile, the hierarchical structure reconstruction task adopts the Semantic-TEDS \cite{Ma2023HRDoc} as its evaluation metric.

\textbf{Comp-HRDoc} is our proposed benchmark, specifically designed for comprehensive hierarchical document structure analysis. It encompasses tasks such as page object detection, reading order prediction, table of contents extraction, and hierarchical structure reconstruction. Comp-HRDoc is built upon the HRDoc-Hard dataset \cite{Ma2023HRDoc}, which comprises 1,000 documents for training and 500 documents for testing. We retain all original images without modification and extend the original annotations to accommodate the evaluation of these included tasks.

In the page object detection task, we utilize the COCO-style segmentation-based mean average precision (mAP) evaluation metric rather than a box-based metric. This choice is due to the fact that a paragraph in HRDoc is considered a logical paragraph, which may span multiple columns in a multi-column page. Consequently, the paragraph's segmentation is derived from the union of segmentations for all its text-lines. Meanwhile, the segmentation of a graphical page object remains identical to its bounding box.
Regarding the reading order prediction task, a document may encompass multiple reading order groups, such as multiple articles in newspapers, without an explicit reading order definition between them. This characteristic aligns with realistic user requirements. However, it renders some previous full ranking metrics \cite{Quiros2022reading} unsuitable for this situation. Furthermore, earlier reading order evaluation metrics primarily focused on the reading order of text units (e.g., text-lines) and neglected paragraph segmentation errors stemming from the Detect stage. As a result, they did not provide a comprehensive assessment of the reading order. In this paper, we propose a reading edit distance score (\textbf{REDS}) to evaluate the reading order task. Specifically, we primarily categorize reading order groups into two types: Text Region Reading Order Group and Graphical Region Reading Order Group, evaluating these two types of reading order groups independently. We define the basic evaluation units as text-lines and graphical page objects. To evaluate the paragraph segmentation error, we introduce a special tag </p> at paragraph ending positions within the reading order groups to serve as a marker for paragraph segmentation. We adopt the Levenshtein distance \cite{levenshtein1966binary}, which measures the minimum number of node operations (insertions, deletions, or substitutions) required to equalize two lists of nodes, to calculate the distance between two reading order groups. Given the presence of multiple groups, for each predicted reading order group, we compute the distance between it and all reading order groups in the ground truth. Subsequently, we utilize the Hungarian matching \cite{kuhn1955hungarian} to obtain the overall minimum distance $D$ and multiply it by the normalization factor $1/N$, where $N$ represents the number of basic units. Ultimately, we define $1-\frac{D}{N}$ as the evaluation score for the reading order prediction task.
For both the table of contents extraction and hierarchical structure reconstruction tasks, we opt for the Semantic-TEDS \cite{Ma2023HRDoc} as their evaluation metric.

\subsection{Implementation Details}

We implement our approach using PyTorch v1.10, and all experiments are conducted on a workstation equipped with 8 Nvidia Tesla V100 GPUs (32 GB memory). It is crucial to mention that in PubLayNet, a list constitutes an entire object containing multiple list items with labels that are inconsistent with those of text or titles. To minimize ambiguity, we treat all lists as specific graphical page objects. 

Since only the task of page object detection needs to be evaluated on PubLayNet and DocLayNet datasets, we only trained the Detect stage in our framework on these two datasets. In our experiments with PubLayNet and DocLayNet, we leverage three multi-scale feature maps $\{C_3, C_4, C_5\}$ from the backbone network, along with the DINO-based graphical page object detection model, to identify graphical objects. In training, the parameters of the CNN backbone network are initialized with a ResNet-50 model \cite{he2016deep} pretrained on the ImageNet classification task, while the parameters of the text embedding extractor are initialized with the pretrained BERT$_{\mathrm{BASE}}$ model \cite{devlin2018bert}. We optimize the models using the AdamW \cite{loshchilov2017decoupled} algorithm with a batch size of 16 and trained for 12 epochs on PubLayNet and 24 epochs on DocLayNet. The learning rate and weight decay are set to 1e-5 and 1e-4 for the CNN backbone network, and 2e-5 and 1e-2 for BERT$_{\mathrm{BASE}}$, respectively. The learning rate is divided by 10 at the $11^{\rm{th}}$ epoch for PubLayNet and $20^{\rm{th}}$ epoch for DocLayNet. Other hyperparameters of AdamW, including betas and epsilon, are set to (0.9, 0.999) and 1e-8, respectively. We also adopt a multi-scale training strategy, randomly rescaling the shorter side of each image to lengths chosen from [512, 640, 768], ensuring the longer side does not exceed 800. During the testing phase, we set the shorter side of the input image to 640.

For HRDoc and Comp-HRDoc, we utilize four multi-scale feature maps $\{C_2, C_3, C_4, C_5\}$ from the backbone network, in conjunction with the Mask2Former-based graphical page object detection model, to identify graphical objects. Given that hierarchical document structure analysis requires processing dozens of document pages, we choose the ResNet-18 model as the CNN backbone network to reduce GPU memory requirements. The parameters of the text embedding extractor are also initialized with the pretrained BERT$_{\mathrm{BASE}}$ model. The models are optimized using the AdamW \cite{loshchilov2017decoupled} algorithm with a batch size of 1 and trained for 20 epochs on HRDoc and Comp-HRDoc. The initial learning rate and weight decay are set to 2e-4 and 1e-2 for the CNN backbone network, and 4e-5 and 1e-2 for BERT$_{\mathrm{BASE}}$, respectively. After a warmup period (set to 2 epochs) during which it increases linearly from 0 to the initial learning rate set in the optimizer, the learning rate linearly decreases from the initial learning rate set in the optimizer to 0. For multi-scale training strategy, the shorter side of each image is randomly rescaled to a length chosen from [320, 416, 512, 608, 704, 800], ensuring that the longer side does not exceed 1024. During the testing phase, we set the shorter side of the input image to 512.

\subsection{Comparisons with Prior Arts}

The Detect module we proposed is a novel combination of top-down and bottom-up approaches for page object detection. Therefore, we first validate the effectiveness of our method on two large-scale document layout analysis datasets, i.e., DocLayNet and PubLayNet.

\textbf{DocLayNet}. We compare our proposed Detect module with the other most competitive methods, including Mask R-CNN, Faster R-CNN, YOLOv5, and DINO on DocLayNet. As shown in Table~\ref{tab-doclaynet-test}, our approach substantially outperforms the closest method YOLOv5 by improving mAP from 76.8\% to 81.0\%. Considering that DocLayNet is an extremely challenging dataset that covers a variety of document scenarios and contains a large number of text regions with fine-grained logical roles, the superior performance achieved by our proposed approach demonstrates the advantage of our approach.
\setlength{\tabcolsep}{4pt}
\begin{table}[ht]
\setlength{\belowcaptionskip}{0.2cm}
\small
\centering
\caption{Performance comparisons on the DocLayNet testing set (in \%). The results of Mask R-CNN, Faster R-CNN, and YOLOv5 are obtained from \cite{pfitzmann2022doclaynet}.}
\label{tab-doclaynet-test}
\begin{tabular}{l|c|ccccc}
\hline & Human & Mask R-CNN & Faster R-CNN & YOLOv5 & DINO & Ours \\
\hline Caption & $\operatorname{84-89}$ & $71.5$ & $70.1$ & $77.7$ & $\textbf{85.5}$ & 83.2\\
Footnote & $\operatorname{83-91}$ & $71.8$ & $73.7$ & $\textbf{77.2}$ & $69.2$ & 69.7\\
Formula & $\operatorname{83-85}$ & $63.4$ & $63.5$ & $\textbf{66.2}$ & $63.8$ & 63.4\\
List-item & $\operatorname{87-88}$ & $80.8$ & $81.0$ & $86.2$ & $80.9$ & \textbf{88.6}\\
Page-footer & $\operatorname{93-94}$ & $59.3$ & $58.9$ & $61.1$ & $54.2$ & \textbf{90.0}\\
Page-header & $\operatorname{85-89}$ & $70.0$ & $72.0$ & $67.9$ & $63.7$ & \textbf{76.3}\\
Picture & $\operatorname{69-71}$ & $72.7$ & $72.0$ & $77.1$ & $\textbf{84.1}$ & 81.6\\
Section-header & $\operatorname{83-84}$ & $69.3$ & $68.4$ & $74.6$ & $64.3$ & \textbf{83.2}\\
Table & $\operatorname{77-81}$ & $82.9$ & $82.2$ & $\textbf{86.3}$ & $85.7$ & 84.8\\
Text & $\operatorname{84-86}$ & $85.8$ & $85.4$ & $\textbf{88.1}$ & $83.3$ & 84.8\\
Title & $\operatorname{60-72}$ & $80.4$ & $79.9$ & $82.7$ & $82.8$ & \textbf{84.9}\\
\hline mAP & $\operatorname{82-83}$ & $73.5$ & $73.4$ & $76.8$ & $74.3$ & \textbf{81.0}\\
\hline
\end{tabular}
\end{table}

\textbf{PubLayNet}. We also compare our approach with several state-of-the-art vision-based and multimodal methods on PubLayNet. The experimental results are presented in Table~\ref{tab-publaynet-val} and Table~\ref{tab-publaynet-test}. We can see that our approach outperforms all these methods regardless of whether textual features are used in our bottom-up text region detection model. 

\setlength{\tabcolsep}{4pt}
\begin{table}[ht]
\setlength{\belowcaptionskip}{0.2cm}
\small
\centering
\caption{Performance comparisons on the PubLayNet validation set (in \%).  Vision and Text stand for using visual and textual features, respectively.}
\label{tab-publaynet-val}
\begin{tabular}{l|l|l|l|l|l|l|l}
\hline Method & Modality& Text & Title & List & Table & Figure & mAP \\
\hline Faster R-CNN \cite{zhong2019publaynet} & Vision & 91.0 & 82.6 & 88.3 & 95.4 & 93.7 & 90.2 \\
Mask R-CNN \cite{zhong2019publaynet} & & 91.6 & 84.0 & 88.6 & 96.0 & 94.9 & 91.0 \\
Naik et al. \cite{naik2022investigating}& & 94.3 & 88.7 & 94.3 & 97.6 & 96.1 & 94.2 \\
Minouei et al. \cite{minouei2021document} & & 94.4 & 90.8 & 94.0 & 97.4 & 96.6 & 94.6 \\
DiT-L \cite{li2022dit} & & 94.4 & 89.3 & 96.0 & 97.8 & \textbf{97.2} & 94.9 \\
SRRV \cite{bi2022srrv} & & 95.8 & 90.1 & 95.0 & 97.6 & 96.7 & 95.0 \\
DINO \cite{zhang2023dino} & & 94.9 & 91.4 & 96.0 & 98.0 & 97.3 & 95.5 \\
TRDLU \cite{yang2022transformer} & & 95.8 & 92.1 & 97.6 & 97.6 & 96.6 & 96.0 \\
\hline
UDoc \cite{gu2022unified} & Vision+Text & 93.9 & 88.5 & 93.7 & 97.3 & 96.4 & 93.9 \\
LayoutLMv3 \cite{huang2022layoutlmv3} & & 94.5 & 90.6 & 95.5 & 97.9 & 97.0 & 95.1 \\
VSR \cite{zhang2021vsr} & & 96.7& 93.1 & 94.7 & 97.4 & 96.4 & 95.7 \\
\hline
Ours & Vision & 97.0 & 92.8 & 96.4 & 98.1 & \textbf{97.4} & 96.3 \\
Ours & Vision+Text & \textbf{97.4} & \textbf{93.5} & \textbf{96.4} & \textbf{98.2} & 97.2 & \textbf{96.5} \\
\hline
\end{tabular}
\end{table}

\setlength{\tabcolsep}{4pt}
\begin{table}[ht]
\setlength{\belowcaptionskip}{0.2cm}
\small
\centering
\caption{Performance comparisons on the PubLayNet test set (in \%).  Vision and Text stand for using visual and textual features, respectively. \label{tab-publaynet-test}}
\begin{tabular}{l|l|l|l|l|l|l|l}
\hline Method & Modality& Text & Title & List & Table & Figure & mAP \\
\hline Faster R-CNN \cite{zhong2019publaynet} & Vision & 91.3 & 81.2 & 88.5 & 94.3 & 94.5 & 90.0 \\
Mask R-CNN \cite{zhong2019publaynet} & & 91.7 & 82.8 & 88.7 & 94.7 & 95.5 & 90.7 \\
DocInsightAI \cite{zhang2021vsr} & & 94.5 & 88.3 & 94.8 & 95.8 & 97.5 & 94.2 \\
SCUT \cite{zhang2021vsr} & & 94.3 & 89.7 & 94.3 & 96.6 & 97.7 & 94.5 \\
SRK \cite{zhang2021vsr} & & 94.7 & 90.0 & 95.1 & 97.2 & \textbf{98.0} & 95.0 \\
SiliconMinds \cite{zhang2021vsr} & & 96.2 & 89.8 & 94.6 & 97.0 & 97.6 & 95.0 \\
\hline
VSR \cite{zhang2021vsr} & Vision+Text & \textbf{96.7}& 92.3 & 94.6 & 97.0 & 97.9 & 95.7 \\
\hline
Ours & Vision & 95.0 & 96.4 & 95.2 & 97.0 & 97.8 & 96.3 \\
Ours & Vision+Text & 95.0 & \textbf{96.6} & \textbf{95.7} & \textbf{97.3} & 97.7 & \textbf{96.5} \\
\hline
\end{tabular}
\end{table}

To further validate the effectiveness of our proposed tree construction based framework for hierarchical document structure analysis, we performed experiments with our method on both HRDoc and Comp-HRDoc datasets and made thorough comparisons with previous approaches.

\textbf{HRDoc}. As demonstrated in Table~\ref{tab-hrdoc-role} and Table~\ref{tab-hrdoc-recon}, we conducted separate performance evaluations for the two tasks in HRDoc, specifically semantic unit classification and hierarchical structure reconstruction. For semantic unit classification, it is evident that our proposed method achieves superior performance in the majority of categories, particularly in the \textit{Fstl} (Firstline) and \textit{Footn} (Footnote) classes, where our approach significantly surpasses previous methods. Although the DSPS Encoder is also a multimodal technique that integrates visual and linguistic information, its performance in the \textit{Mail} category is notably inferior to that of Sentence-BERT. However, on HRDoc-Hard, our method attains an F1 score nearly 5\% higher than the DSPS Encoder in this category.
Regarding hierarchical structure reconstruction, our proposed tree construction based method markedly outperforms the DSPS Encoder. On HRDoc-Hard, we exceed its performance by 16.63\% and 15.77\% in Micro-STEDS and Macro-STEDS, respectively. Similarly, on HRDoc-Simple, we surpass the DSPS Encoder by 13.61\% and 13.36\% in Micro-STEDS and Macro-STEDS, respectively. It is important to highlight that our proposed method evaluates the performance based on the predicted reading order sequence, whereas the DSPS Encoder directly takes advantage of the ground-truth reading order.

\setlength{\tabcolsep}{2pt}
\begin{table}[ht]
\setlength{\belowcaptionskip}{0.2cm}
\small
\centering
\caption{Comparison results of different baseline models in the semantic unit classification task on HRDoc (in \%). F1 means F1-score. The results of Cascade-RCNN, ResNet+RoIAlign, Sentence-Bert and DSPS Encoder are all obtained from \cite{Ma2023HRDoc}.}
\label{tab-hrdoc-role}
\begin{adjustbox}{width=\textwidth}
\begin{tabular}{c|cccccccccccccccc}
\hline
\multirow{3}{*}{Method}                                          & \multicolumn{16}{c}{HRDoc-Hard  F1 (\%)}                                      \\ \cline{2-17} 
& \multicolumn{1}{c|}{\multirow{2}{*}{Title}} & \multicolumn{1}{c|}{\multirow{2}{*}{Author}} & \multicolumn{1}{c|}{\multirow{2}{*}{Mail}} & \multicolumn{1}{c|}{\multirow{2}{*}{Affili}} & \multicolumn{1}{c|}{\multirow{2}{*}{Sect}} & \multicolumn{1}{c|}{\multirow{2}{*}{Fstl}} & \multicolumn{1}{c|}{\multirow{2}{*}{Paral}} & \multicolumn{1}{c|}{\multirow{2}{*}{Table}} & \multicolumn{1}{c|}{\multirow{2}{*}{Fig}} & \multicolumn{1}{c|}{\multirow{2}{*}{Cap}} & \multicolumn{1}{c|}{\multirow{2}{*}{Equ}} & \multicolumn{1}{c|}{\multirow{2}{*}{Foot}} & \multicolumn{1}{c|}{\multirow{2}{*}{Head}} & \multicolumn{1}{c|}{\multirow{2}{*}{Footn}} & \multicolumn{2}{c}{Avg. F1 (\%)}   \\ \cline{16-17} 
& \multicolumn{1}{c|}{}                       & \multicolumn{1}{c|}{}                        & \multicolumn{1}{c|}{}                      & \multicolumn{1}{c|}{}                        & \multicolumn{1}{c|}{}                      & \multicolumn{1}{c|}{}                      & \multicolumn{1}{c|}{}                       & \multicolumn{1}{c|}{}                       & \multicolumn{1}{c|}{}                     & \multicolumn{1}{c|}{}                     & \multicolumn{1}{c|}{}                     & \multicolumn{1}{c|}{}                      & \multicolumn{1}{c|}{}                      & \multicolumn{1}{c|}{}                       & \multicolumn{1}{c|}{Micro} & Macro \\ \hline
Cascade-RCNN                                                     & \multicolumn{1}{c|}{81.50}                  & \multicolumn{1}{c|}{49.77}                   & \multicolumn{1}{c|}{33.39}                 & \multicolumn{1}{c|}{49.34}                   & \multicolumn{1}{c|}{75.92}                 & \multicolumn{1}{c|}{64.96}                 & \multicolumn{1}{c|}{77.86}                  & \multicolumn{1}{c|}{69.96}                  & \multicolumn{1}{c|}{72.22}                & \multicolumn{1}{c|}{43.72}                & \multicolumn{1}{c|}{68.84}                & \multicolumn{1}{c|}{70.91}                 & \multicolumn{1}{c|}{71.00}                 & \multicolumn{1}{c|}{52.67}                  & \multicolumn{1}{c|}{73.37} & 64.94 \\ \hline
\begin{tabular}[c]{@{}c@{}}ResNet+RoIAlign\end{tabular} & \multicolumn{1}{c|}{82.40}                  & \multicolumn{1}{c|}{48.40}                   & \multicolumn{1}{c|}{18.43}                 & \multicolumn{1}{c|}{61.33}                   & \multicolumn{1}{c|}{33.66}                 & \multicolumn{1}{c|}{45.37}                 & \multicolumn{1}{c|}{87.99}                  & \multicolumn{1}{c|}{21.89}                  & \multicolumn{1}{c|}{70.28}                & \multicolumn{1}{c|}{61.54}                & \multicolumn{1}{c|}{48.32}                & \multicolumn{1}{c|}{73.69}                 & \multicolumn{1}{c|}{75.71}                 & \multicolumn{1}{c|}{6.79}                   & \multicolumn{1}{c|}{79.25} & 52.56 \\ \hline
Sentence-Bert                                                    & \multicolumn{1}{c|}{95.85}                  & \multicolumn{1}{c|}{89.92}                   & \multicolumn{1}{c|}{\textbf{91.68}}                 & \multicolumn{1}{c|}{\textbf{91.75}}                   & \multicolumn{1}{c|}{94.26}                 & \multicolumn{1}{c|}{88.68}                 & \multicolumn{1}{c|}{96.77}                  & \multicolumn{1}{c|}{76.96}                  & \multicolumn{1}{c|}{91.67}                & \multicolumn{1}{c|}{91.99}                & \multicolumn{1}{c|}{93.94}                & \multicolumn{1}{c|}{94.68}                 & \multicolumn{1}{c|}{92.65}                 & \multicolumn{1}{c|}{62.61}                  & \multicolumn{1}{c|}{94.68} & 89.53 \\ \hline
DSPS Encoder                                                     & \multicolumn{1}{c|}{\textbf{97.71}}                  & \multicolumn{1}{c|}{93.93}                   & \multicolumn{1}{c|}{85.49}                 & \multicolumn{1}{c|}{90.95}                   & \multicolumn{1}{c|}{96.06}                 & \multicolumn{1}{c|}{91.24}                 & \multicolumn{1}{c|}{97.96}                  & \multicolumn{1}{c|}{\textbf{100.0}}                  & \multicolumn{1}{c|}{\textbf{100.0}}                & \multicolumn{1}{c|}{\textbf{97.32}}                & \multicolumn{1}{c|}{\textbf{97.92}}                & \multicolumn{1}{c|}{98.54}                 & \multicolumn{1}{c|}{\textbf{97.83}}                 & \multicolumn{1}{c|}{88.84}                  & \multicolumn{1}{c|}{96.74} & 95.27 \\ \hline
Ours                                                             & \multicolumn{1}{c|}{97.26}                  & \multicolumn{1}{c|}{\textbf{94.22}}                   & \multicolumn{1}{c|}{90.33}                 & \multicolumn{1}{c|}{90.73}                   & \multicolumn{1}{c|}{\textbf{96.25}}                 & \multicolumn{1}{c|}{\textbf{94.09}}        & \multicolumn{1}{c|}{\textbf{98.55}}                  & \multicolumn{1}{c|}{\textbf{100.0}}                  & \multicolumn{1}{c|}{\textbf{100.0}}                & \multicolumn{1}{c|}{96.41}                & \multicolumn{1}{c|}{97.68}                & \multicolumn{1}{c|}{\textbf{98.57}}                 & \multicolumn{1}{c|}{97.79}                 & \multicolumn{1}{c|}{\textbf{90.75}}                  & \multicolumn{1}{c|}{\textbf{97.59}} & \textbf{95.90} \\ \hline\hline
\multirow{3}{*}{Method}                                          & \multicolumn{16}{c}{HRDoc-Simple F1 (\%)}                                            \\ \cline{2-17} 
& \multicolumn{1}{c|}{\multirow{2}{*}{Title}} & \multicolumn{1}{c|}{\multirow{2}{*}{Author}} & \multicolumn{1}{c|}{\multirow{2}{*}{Mail}} & \multicolumn{1}{c|}{\multirow{2}{*}{Affili}} & \multicolumn{1}{c|}{\multirow{2}{*}{Sect}} & \multicolumn{1}{c|}{\multirow{2}{*}{Fstl}} & \multicolumn{1}{c|}{\multirow{2}{*}{Paral}} & \multicolumn{1}{c|}{\multirow{2}{*}{Table}} & \multicolumn{1}{c|}{\multirow{2}{*}{Fig}} & \multicolumn{1}{c|}{\multirow{2}{*}{Cap}} & \multicolumn{1}{c|}{\multirow{2}{*}{Equ}} & \multicolumn{1}{c|}{\multirow{2}{*}{Foot}} & \multicolumn{1}{c|}{\multirow{2}{*}{Head}} & \multicolumn{1}{c|}{\multirow{2}{*}{Footn}} & \multicolumn{2}{c}{Avg. F1 (\%)}   \\ \cline{16-17} 
& \multicolumn{1}{c|}{}                       & \multicolumn{1}{c|}{}                        & \multicolumn{1}{c|}{}                      & \multicolumn{1}{c|}{}                        & \multicolumn{1}{c|}{}                      & \multicolumn{1}{c|}{}                      & \multicolumn{1}{c|}{}                       & \multicolumn{1}{c|}{}                       & \multicolumn{1}{c|}{}                     & \multicolumn{1}{c|}{}                     & \multicolumn{1}{c|}{}                     & \multicolumn{1}{c|}{}                      & \multicolumn{1}{c|}{}                      & \multicolumn{1}{c|}{}                       & \multicolumn{1}{c|}{Micro} & Macro \\ \hline
Cascade-RCNN                                                     & \multicolumn{1}{c|}{78.83}                  & \multicolumn{1}{c|}{72.74}                   & \multicolumn{1}{c|}{64.54}                 & \multicolumn{1}{c|}{70.13}                   & \multicolumn{1}{c|}{91.35}                 & \multicolumn{1}{c|}{87.53}                 & \multicolumn{1}{c|}{89.7}                   & \multicolumn{1}{c|}{89.3}                   & \multicolumn{1}{c|}{73.87}                & \multicolumn{1}{c|}{64.87}                & \multicolumn{1}{c|}{83.87}                & \multicolumn{1}{c|}{87.5}                  & \multicolumn{1}{c|}{-}                     & \multicolumn{1}{c|}{79.32}                  & \multicolumn{1}{c|}{88.30} & 80.85 \\ \hline
\begin{tabular}[c]{@{}c@{}}ResNet+RoIAlign\end{tabular} & \multicolumn{1}{c|}{93.67}                  & \multicolumn{1}{c|}{82.53}                   & \multicolumn{1}{c|}{81.33}                 & \multicolumn{1}{c|}{84.39}                   & \multicolumn{1}{c|}{37.09}                 & \multicolumn{1}{c|}{38.39}                 & \multicolumn{1}{c|}{91.86}                  & \multicolumn{1}{c|}{58.44}                  & \multicolumn{1}{c|}{48.53}                & \multicolumn{1}{c|}{70.75}                & \multicolumn{1}{c|}{26.89}                & \multicolumn{1}{c|}{98.33}                 & \multicolumn{1}{c|}{-}                     & \multicolumn{1}{c|}{49.76}                  & \multicolumn{1}{c|}{85.61} & 66.30 \\ \hline
Sentence-Bert                                                    & \multicolumn{1}{c|}{98.98}                  & \multicolumn{1}{c|}{96.47}                   & \multicolumn{1}{c|}{\textbf{98.95}}                 & \multicolumn{1}{c|}{97.42}                   & \multicolumn{1}{c|}{97.3}                  & \multicolumn{1}{c|}{93.27}                 & \multicolumn{1}{c|}{98.72}                  & \multicolumn{1}{c|}{94.42}                  & \multicolumn{1}{c|}{95.72}                & \multicolumn{1}{c|}{93.36}                & \multicolumn{1}{c|}{96.02}                & \multicolumn{1}{c|}{99.89}                 & \multicolumn{1}{c|}{-}                     & \multicolumn{1}{c|}{87.11}                  & \multicolumn{1}{c|}{97.74} & 95.97 \\ \hline
DSPS Encoder                                                     & \multicolumn{1}{c|}{99.43}                  & \multicolumn{1}{c|}{98.83}                   & \multicolumn{1}{c|}{96.45}                 & \multicolumn{1}{c|}{97.33}                   & \multicolumn{1}{c|}{\textbf{99.6}}                  & \multicolumn{1}{c|}{98.22}                 & \multicolumn{1}{c|}{\textbf{99.74}}                  & \multicolumn{1}{c|}{\textbf{100.0}}                    & \multicolumn{1}{c|}{99.95}                & \multicolumn{1}{c|}{\textbf{99.06}}                & \multicolumn{1}{c|}{\textbf{97.91}}                & \multicolumn{1}{c|}{\textbf{100.0}}                   & \multicolumn{1}{c|}{-}                     & \multicolumn{1}{c|}{99.15}                  & \multicolumn{1}{c|}{99.52} & 98.90 \\ \hline
Ours                                                             & \multicolumn{1}{c|}{\textbf{99.67}}                  & \multicolumn{1}{c|}{\textbf{98.98}}                   & \multicolumn{1}{c|}{98.78}                 & \multicolumn{1}{c|}{\textbf{98.95}}                   & \multicolumn{1}{c|}{99.39}                 & \multicolumn{1}{c|}{\textbf{98.51}}                 & \multicolumn{1}{c|}{\textbf{99.74}}                  & \multicolumn{1}{c|}{\textbf{100.0}}                    & \multicolumn{1}{c|}{\textbf{100.0}}                  & \multicolumn{1}{c|}{98.03}                & \multicolumn{1}{c|}{97.07}                & \multicolumn{1}{c|}{\textbf{100.0}}                   & \multicolumn{1}{c|}{-}                     & \multicolumn{1}{c|}{\textbf{99.57}}                  & \multicolumn{1}{c|}{\textbf{99.54}} & \textbf{99.13} \\ \hline
\end{tabular}
\end{adjustbox}
\end{table}

\setlength{\tabcolsep}{4pt}
\begin{table}[ht]
\setlength{\belowcaptionskip}{0.2cm}
\small
\centering
\caption{Comparison results of different models in the hierarchical document reconstruction task on HRDoc.}
\label{tab-hrdoc-recon}
\begin{tabular}{c|c|cc|cc}
\hline
\multirow{2}{*}{Method} & \multirow{2}{*}{Level} & \multicolumn{2}{c|}{HRDoc-Simple}                                   & \multicolumn{2}{c}{HRDoc-Hard}                                     \\ \cline{3-6} 
                        &                        & \multicolumn{1}{c|}{Micro-STEDS} & \multicolumn{1}{l|}{Macro-STEDS} & \multicolumn{1}{c|}{Micro-STEDS} & \multicolumn{1}{l}{Macro-STEDS} \\ \hline
DocParser               & Page                   & \multicolumn{1}{c|}{0.2361}      & 0.2506                           & \multicolumn{1}{c|}{0.1873}      & 0.2015                          \\ \hline
DSPS Encoder                   & Document               & \multicolumn{1}{c|}{0.8143}      & 0.8174                           & \multicolumn{1}{c|}{0.6903}      & 0.6971                          \\ \hline
Ours                    & Document               & \multicolumn{1}{c|}{\textbf{0.9504}}      & \textbf{0.9510 }                          & \multicolumn{1}{c|}{\textbf{0.8566}}      & \textbf{0.8548}                          \\ \hline
\end{tabular}
\end{table}

\textbf{Comp-HRDoc}. As presented in Table~\ref{tab-comp-hrdoc}, we conduct a comprehensive performance evaluation for all tasks in Comp-HRDoc, encompassing page object detection, reading order prediction, table of contents extraction, and hierarchical document reconstruction. We select previous state-of-the-art methods specifically designed for each task to be evaluated using our benchmark. Our proposed method is capable of handling all tasks concurrently and achieves significantly superior results in each of them. Specifically, for page object detection, our method surpasses Mask2former \cite{cheng2022mask2former} by 14.52\% in terms of segmentation-based mAP. Regarding reading order prediction, as previous methods rarely consider multiple reading order groups, we have enhanced the partial order-based algorithm proposed by Lorenzo et al. \cite{Quiros2022reading} to decode both categories of reading order groups simultaneously. We observe that in the more challenging category (i.e., text region reading order group), our method outperforms their approach by 15.78\% in terms of previously defined REDS. For table of contents extraction, our method exceeds the Multimodal Tree Decoder (MTD) \cite{hu2022toc} by 18.50\% and 16.87\% in Micro-STEDS and Macro-STEDS, respectively. In hierarchical structure reconstruction, the evaluation of the DSPS Encoder \cite{Ma2023HRDoc} depends on the provided reading order ground-truth and bounding box ground-truth for graphical objects, while our method's result is entirely independent of ground truth and is obtained through a comprehensive end-to-end evaluation. Under these conditions, our method still surpasses the DSPS Encoder by 14.68\% and 13.94\% in Micro-STEDS and Macro-STEDS, respectively. Because the Comp-HRDoc benchmark supports a holistic end-to-end evaluation process for hierarchical document structure analysis, it offers a better evaluation benchmark for universal layout analysis.

\setlength{\tabcolsep}{4pt}
\begin{table}[ht]
\setlength{\belowcaptionskip}{0.2cm}
\small
\centering
\caption{Comparison results of different models in tasks including page object detection, reading order prediction, table of contents extraction and hierarchical document reconstruction on Comp-HRDoc. The symbol $^\dag$ represents the results of our enhanced replication, whereas $^\ddag$ indicates that the evaluation of this result relies on the provided reading order ground-truth and bounding box ground-truth for graphical objects.}
\label{tab-comp-hrdoc}
\begin{adjustbox}{width=\textwidth}
\begin{tabular}{c|c|cc|cc|cc}
\hline
\multirow{2}{*}{Methods} & Page Object Detection                                    & \multicolumn{2}{c|}{Reading Order Prediction}                                                                                                                                               & \multicolumn{2}{c|}{Table of Contents Extraction}      & \multicolumn{2}{c}{Hierarchical Reconstruction}        \\ \cline{2-8} 
                         & \begin{tabular}[c]{@{}c@{}}Segmentation\\ mAP (\%) \end{tabular} & \multicolumn{1}{c|}{\begin{tabular}[c]{@{}c@{}}Text Region\\ REDS\end{tabular}} & \begin{tabular}[c]{@{}c@{}}Graphical Region\\ REDS\end{tabular} & \multicolumn{1}{c|}{Micro-STEDS}     & Macro-STEDS     & \multicolumn{1}{c|}{Micro-STEDS}     & Macro-STEDS     \\ \hline
Mask2Former \cite{cheng2022mask2former}              & 73.54                                                    & \multicolumn{1}{c|}{-}                                                                               & -                                                                                   & \multicolumn{1}{c|}{-}               & -               & \multicolumn{1}{c|}{-}               & -               \\ \hline
Lorenzo et al.$^\dag$ \cite{Quiros2022reading}          & -                                                        & \multicolumn{1}{c|}{0.7741}                                                                          & 0.8583                                                                              & \multicolumn{1}{c|}{-}               & -               & \multicolumn{1}{c|}{-}               & -               \\ \hline
MTD \cite{hu2022toc}                      & -                                                        & \multicolumn{1}{c|}{-}                                                                               & -                                                                                   & \multicolumn{1}{c|}{0.6755}          & 0.7099          & \multicolumn{1}{c|}{-}               & -               \\ \hline
DSPS Encoder$^\ddag$ \cite{Ma2023HRDoc}             & -                                                        & \multicolumn{1}{c|}{-}                                                                               & -                                                                                   & \multicolumn{1}{c|}{0.5754}          & 0.6230          & \multicolumn{1}{c|}{0.6903}          & 0.6971          \\ \hline
Ours                     & \textbf{88.06}                                           & \multicolumn{1}{c|}{\textbf{0.9319}}                                                                 & \textbf{0.8637}                                                                     & \multicolumn{1}{c|}{\textbf{0.8605}} & \textbf{0.8788} & \multicolumn{1}{c|}{\textbf{0.8371}} & \textbf{0.8365} \\ \hline
\end{tabular}
\end{adjustbox}
\end{table}

\subsection{Ablation Studies}

We conducted a series of ablation experiments based on Comp-HRDoc to verify the impact of using different modules and modalities.

\textbf{Effectiveness of the hybrid strategy and multimodality in the Detect module.}
In this section, we first evaluate the effectiveness of the proposed hybrid strategy in the Detect module. To this end, we train two baseline models: 1) a Mask2Former baseline to detect both graphical page objects and text regions and 2) a hybrid model (denoted as Hybrid (V)) that leverages Mask2Former for graphical object detection and only uses visual and 2D position features for bottom-up text region detection. As shown in the first two rows of Table~\ref{tab-ablation-group-hybrid}, compared with the Mask2Former-R50 model, the Hybrid-R18 (V) model can achieve comparable graphical page object detection results but much higher text region detection accuracy on Comp-HRDoc, leading to a 9.86\% improvement in terms of segmentation-based mAP. In particular, the Hybrid-R18 (V) model can significantly improve small-scale text region detection performance, e.g., 84.67\% vs. 68.97\% for Page-footnote, 95.08\% vs. 59.01\% for Page-header and 95.93\% vs. 62.68\% for Page-footer. These experimental results clearly demonstrate the effectiveness of the proposed hybrid strategy that combines the best of both top-down and bottom-up methods. In addition, we also conducted an ablation experiment to explore the effectiveness of text modalities in the Detect module, as depicted in the last two rows of Table~\ref{tab-ablation-group-hybrid}. We find that the hybrid model with text modality (denoted as Hybrid (V+T)) achieves much better performance in semantically sensitive categories, such as Author, Mail, and Affiliate, leading to a 4.66\% improvement in terms of segmentation-based mAP. Notably, we have observed many cases of inconsistent paragraph annotations in HRDoc, which might be one of the reasons for the relatively lower performance in the \textit{Para} (Paragraph) category.
More ablation studies in the Detect module can be found in our conference paper \cite{zhong2023hybrid}.

\setlength{\tabcolsep}{4pt}
\begin{table}[ht]
\setlength{\belowcaptionskip}{0.2cm}
\small
\centering
\caption{Ablation studies of hybrid strategy and multimodality in the Detect module on Comp-HRDoc (in \%).}
\label{tab-ablation-group-hybrid}
\begin{adjustbox}{width=\textwidth}
\begin{tabular}{c|ccccccccccccc}
\hline
\multirow{2}{*}{Method} & \multicolumn{13}{c}{Page Object Detection}                                                                                                                                                                                                                                                                                                                                              \\ \cline{2-14} 
                        & \multicolumn{1}{c|}{Title} & \multicolumn{1}{c|}{Author} & \multicolumn{1}{c|}{Mail}  & \multicolumn{1}{c|}{Affili} & \multicolumn{1}{c|}{Sect}  & \multicolumn{1}{c|}{Para}  & \multicolumn{1}{c|}{Table} & \multicolumn{1}{c|}{Fig}   & \multicolumn{1}{c|}{Cap}   & \multicolumn{1}{c|}{Foot}  & \multicolumn{1}{c|}{Head}  & \multicolumn{1}{c|}{Footn} & \begin{tabular}[c]{@{}c@{}}Seg \\    mAP\end{tabular} \\ \hline
Mask2Former-R18 \cite{cheng2022mask2former}             & \multicolumn{1}{c|}{84.41} & \multicolumn{1}{c|}{70.92}  & \multicolumn{1}{c|}{51.95} & \multicolumn{1}{c|}{62.13}  & \multicolumn{1}{c|}{76.51} & \multicolumn{1}{c|}{71.31} & \multicolumn{1}{c|}{79.78} & \multicolumn{1}{c|}{86.23} & \multicolumn{1}{c|}{71.77} & \multicolumn{1}{c|}{56.73} & \multicolumn{1}{c|}{54.89} & \multicolumn{1}{c|}{66.50} & 69.42                                                \\ \hline
Mask2Former-R50 \cite{cheng2022mask2former}             & \multicolumn{1}{c|}{85.37} & \multicolumn{1}{c|}{74.67}  & \multicolumn{1}{c|}{66.66} & \multicolumn{1}{c|}{69.21}  & \multicolumn{1}{c|}{78.18} & \multicolumn{1}{c|}{74.62} & \multicolumn{1}{c|}{80.49} & \multicolumn{1}{c|}{86.55} & \multicolumn{1}{c|}{76.10} & \multicolumn{1}{c|}{62.68} & \multicolumn{1}{c|}{59.01} & \multicolumn{1}{c|}{68.97} & 73.54                                                 \\ \hline
Hybrid-R18 (V)      & \multicolumn{1}{c|}{94.50} & \multicolumn{1}{c|}{79.40}  & \multicolumn{1}{c|}{51.57} & \multicolumn{1}{c|}{70.88}  & \multicolumn{1}{c|}{\textbf{89.65}} & \multicolumn{1}{c|}{\textbf{83.39}} & \multicolumn{1}{c|}{80.32} & \multicolumn{1}{c|}{86.35} & \multicolumn{1}{c|}{89.02} & \multicolumn{1}{c|}{95.93} & \multicolumn{1}{c|}{\textbf{95.08}} & \multicolumn{1}{c|}{\textbf{84.67}} & 83.40                                                 \\ \hline
Hybrid-R18 (V + T)    & \multicolumn{1}{c|}{\textbf{94.60}} & \multicolumn{1}{c|}{\textbf{90.66}}  & \multicolumn{1}{c|}{\textbf{84.54}} & \multicolumn{1}{c|}{\textbf{84.91}}  & \multicolumn{1}{c|}{89.48} & \multicolumn{1}{c|}{81.72} & \multicolumn{1}{c|}{\textbf{80.58}} & \multicolumn{1}{c|}{\textbf{86.71}} & \multicolumn{1}{c|}{\textbf{89.40}} & \multicolumn{1}{c|}{\textbf{96.44}} & \multicolumn{1}{c|}{94.96} & \multicolumn{1}{c|}{82.91} & \textbf{88.06}                                               \\ \hline
\end{tabular}
\end{adjustbox}
\end{table}

\textbf{Effectiveness of multimodality in the Construct module.} In this study, we conducted an ablation experiment to investigate the effects of different modalities, specifically text and image modalities. To study the impact of section numbers on the task of table of contents extraction, we removed the section numbers from the text content of section headings and examined the resulting influence on the extraction process, as shown in Table~\ref{tab-ablation-toc-text}. The experiment yields several significant insights.
Firstly, the presence or absence of section numbers in the text content considerably affects the performance when considering the text modality. This observation highlights the pronounced relationship between text modality and section numbers in the extraction of tables of contents.
Secondly, the findings indicate that the image modality, when used independently, performs admirably, achieving higher scores than relying solely on text modality. This demonstrates the robustness of the image modality.
Lastly, the most favorable performance is achieved when both text and image modalities are incorporated into the methodology. This outcome underlines the necessity of employing a multimodal strategy to accomplish the most desirable results in extracting tables of contents.

\setlength{\tabcolsep}{4pt}
\begin{table}[ht]
\setlength{\belowcaptionskip}{0.2cm}
\small
\centering
\caption{Ablation studies of various modalities in the Construct module on Comp-HRDoc.}
\label{tab-ablation-toc-text}
\begin{tabular}{c|c|c|c|c}
\hline
\multicolumn{3}{c|}{Modality}                           & \multirow{3}{*}{Micro-STEDS} & \multirow{3}{*}{Macro-STEDS} \\ \cline{1-3}
\multicolumn{2}{c|}{Text}   & \multirow{2}{*}{Image}    &                              &      \\ \cline{1-2}
w/o Section Number & with Section Number    &            &                              &  \\ \hline
\checkmark &  & &  0.6409            & 0.6834 \\
 & \checkmark & &  0.8341            & 0.8528 \\
 &  & \checkmark &  0.8477            & 0.8685 \\ \hline
 \checkmark & & \checkmark & 0.8436            & 0.8640 \\
 & \checkmark & \checkmark & \textbf{0.8605}            & \textbf{0.8788}  \\ \hline
\end{tabular}
\end{table}

\textbf{Effectiveness of various components in TOC Relation Prediction Head.} 
In this study, we perform an ablation experiment to comparatively assess the influence of individual components within our proposed TOC relation prediction head. The experimental results are presented in Table~\ref{tab-ablation-toc-head}. Firstly, when the Relation Prediction Head for Sibling Finding presented in Fig.~\ref{fig:toc_head} is removed, the performance slightly diminishes to 0.8545 and 0.8712 for Micro-STEDS and Macro-STEDS, respectively. Differently, a substantial performance decline is observed when the Tree Insert Algorithm (i.e., Algorithm \ref{alg:insert}) is omitted, with Micro-STEDS and Macro-STEDS scores of 0.7111 and 0.7652, respectively. Lastly, replacing the softmax cross-entropy loss with the standard binary cross-entropy loss also leads to a decrease in performance, with Micro-STEDS and Macro-STEDS scores of 0.7002 and 0.7475, respectively. These experimental results suggest that each component of our approach positively contributes to the overall performance. Especially, the Tree Insert Algorithm plays a critical role in enhancing the performance of the TOC extraction task.

\setlength{\tabcolsep}{4pt}
\begin{table}[ht]
\setlength{\belowcaptionskip}{0.2cm}
\small
\centering
\caption{Ablation studies of various components in TOC Relation Prediction Head on Comp-HRDoc.}
\label{tab-ablation-toc-head}
\begin{tabular}{l|c|cc}
\hline
\multirow{2}{*}{Method}      & \multirow{2}{*}{Level} & \multicolumn{2}{c}{Table of Contents Extraction}         \\ \cline{3-4} 
                             &                        & \multicolumn{1}{c|}{Micro-STEDS}      & Macro-STEDS      \\ \hline
Ours                         & Document               & \multicolumn{1}{c|}{\textbf{0.8605}} & \textbf{0.8788} \\ \hline
- Sibling Finding            & Document               & \multicolumn{1}{c|}{0.8545}           & 0.8712           \\
\quad - Tree Insert Algorithm      & Document               & \multicolumn{1}{c|}{0.7111}           & 0.7652           \\
\quad \quad - Softmax Cross Entropy Loss & Document               & \multicolumn{1}{c|}{0.7002}           & 0.7475           \\ \hline
\end{tabular}
\end{table}

\begin{figure}[t]
    \centering
    \subfigure[Failure case due to incorrect recognition of section headings.]{
        \includegraphics[width=7cm]{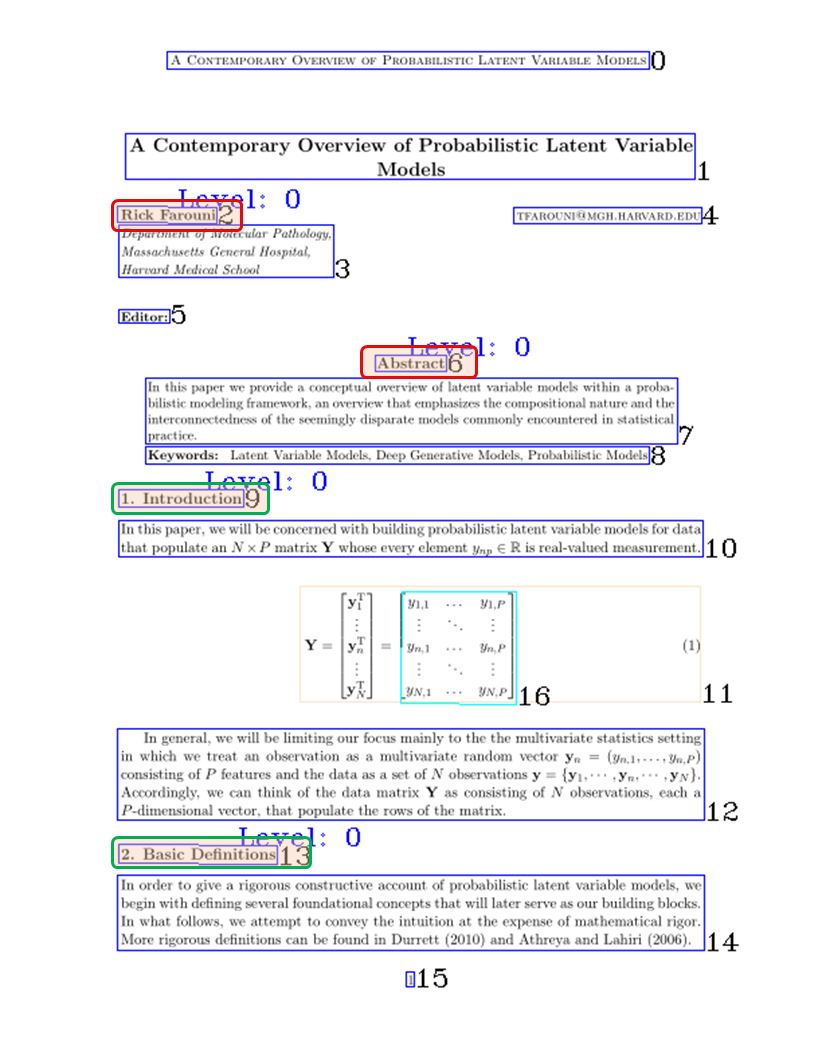}
    }
    \subfigure[Failure case due to the lack of section number.]{
        \includegraphics[width=7cm]{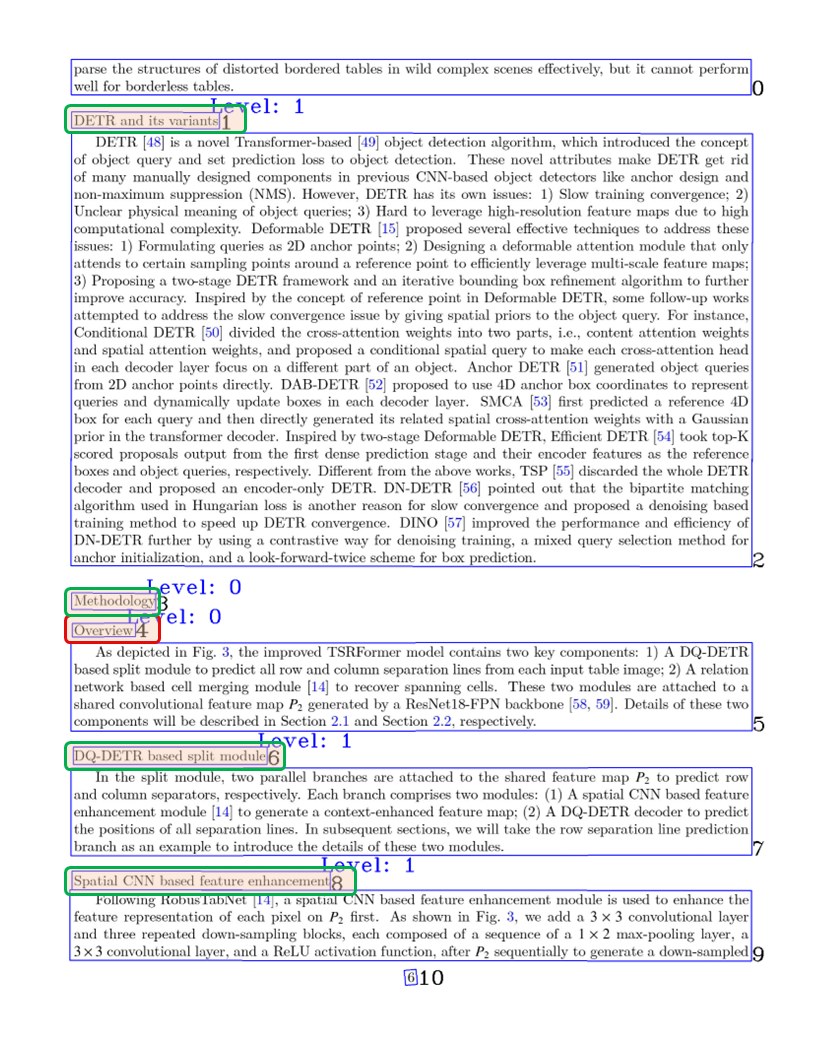}
    }
    \centering
    \caption{Some typical failure cases of Table of Contents extraction.}
    \label{fig:failure_cases}
\end{figure}

\subsection{Limitations of Our Approach}

While our proposed end-to-end system demonstrates outstanding performance in a majority of tasks, as corroborated by prior experiments, it is not without limitations. For instance, we presume that the section headers supplied to the Construct module from previous stages are accurately recognized. Consequently, the recognition performance of section headings accounts for part of the Construct module's bottleneck. Moreover, the information regarding section numbers is vital for harnessing the semantics of section headings within our proposed system. Therefore, for documents lacking section numbers, our approach may not exhibit adequate robustness. Several failure examples are depicted in Fig.~\ref{fig:failure_cases}, with red boxes indicating incorrect predictions and green boxes signifying correct predictions. Note that these difficulties are common challenges faced by other state-of-the-art methods. Finding practical solutions to these problems will be the focus of our future work.
\section{Conclusion and Future Work}

In this study, we perform a thorough examination of various aspects of hierarchical document structure analysis (HDSA) and propose a tree construction based approach, named Detect-Order-Construct, to simultaneously address multiple crucial subtasks in HDSA. To showcase the effectiveness of this novel framework, we design an effective end-to-end solution and uniformly define the tasks of these three stages as relation prediction problems. Moreover, to comprehensively assess the performance of different approaches, we introduce a new benchmark, termed Comp-HRDoc, which concurrently evaluates page object detection, reading order prediction, table of contents extraction, and hierarchical structure reconstruction. As a result, our proposed end-to-end system attains state-of-the-art performance on two large-scale document layout analysis datasets (i.e., PubLayNet and DocLayNet), a hierarchical document structure reconstruction dataset (i.e., HRDoc), and our comprehensive benchmark (i.e., Comp-HRDoc).

In future research, we aim to broaden the scope of our framework to encompass a wider range of real-life scenarios, including contracts, financial reports, and handwritten documents. Additionally, we recognize the importance of addressing documents with graph-based logical structures for more general applications. As such, we plan to explore more robust and effective approaches to handle these complex scenarios. Our ongoing efforts are dedicated to finding a comprehensive and universal document structure analysis solution.



 \bibliographystyle{elsarticle-num} 
 \biboptions{numbers,sort&compress}
 \bibliography{bibfile}





\end{document}